\documentclass[letterpaper, 10 pt, conference]{ieeeconf}  % Comment this line out if you need a4paper

\IEEEoverridecommandlockouts                              % This command is only needed if
\usepackage{graphicx} %include graphics

\usepackage{tabularx} %for tables
\usepackage{booktabs}
\usepackage{caption}
\usepackage{subcaption}
\usepackage{multicol}
\usepackage{tikz}
\usepackage{tikz-3dplot}
\usepackage{arydshln}
\usetikzlibrary{positioning,calc,patterns,angles,quotes}
\usetikzlibrary{quotes,angles}
\usepackage[space]{cite}

\usepackage{xcolor}

\title{\LARGE \bf Mono Video-Based AI Corridor for Model-Free Detection of Collision-Relevant Obstacles}
\author{Thomas Michalke$^{1}$, Yassin Kaddar$^{1*}$, Thomas N\"urnberg$^{1}$, Linh K\"astner$^{2}$, Jens Lambrecht$^{2}$% <-this % stops a space
	\thanks{$^{1}$Robert Bosch GmbH,
		Robert-Bosch-Platz 1,
		70839 Gerlingen-Schillerh\"ohe, Germany
		{\tt\small  Firstname.Lastname@de.bosch.com}}%
	\thanks{$^{2}$Technische Universit\"at Berlin}
	\thanks{{\tt\small  doan.hl.kaestner@campus.tu-berlin.de}}%
	\thanks{{\tt\small  lambrecht@tu-berlin.de}}%
	\thanks{*contact author}
}
\begin{document}
	\maketitle
	\thispagestyle{empty}
	\pagestyle{empty}
	%
	%%%%%%%%%%%%%%%%%%%%%%%%%%%%%%%%%%%%%%%%%%%%%%%%%%%%%%%%%%%%%%%%%%%%%%%%%%%%%%%%
	\begin{abstract}
		The detection of previously unseen, unexpected obstacles on the road is a major challenge for automated driving systems. Different from the detection of ordinary objects with pre-definable classes, detecting unexpected obstacles on the road cannot be resolved by upscaling the sensor technology alone (e.g., high resolution video imagers / radar antennas, denser LiDAR scan lines). This is due to the
		fact, that there is a wide variety in the types of unexpected obstacles that also do not share a common appearance (e.g., lost cargo as a suitcase or bicycle, tire fragments, a tree stem). Also adding object classes or adding ``all'' of these objects to a common ``unexpected obstacle'' class does not scale.
		In this contribution, we study the feasibility of using a deep learning video-based lane corridor (called ``AI ego-corridor'') to ease the challenge by inverting the problem: Instead of detecting a previously unseen object, the AI ego-corridor detects that the ego-lane ahead ends. A smart ground-truth definition enables an easy feature-based classification of an abrupt end of the ego-lane.
		We propose two neural network designs and research among other things the potential of training with synthetic data.
		We evaluate our approach on a test vehicle platform. It is shown that the approach is able to detect numerous previously unseen obstacles at a distance of up to 300\,m with a detection rate of 95\,\%.
	\end{abstract}
	%
	%%%%%%%%%%%%%%%%%%%%%%%%%%%%%%%%%%%%%%%%%%%%%%%%%%%%%%%%%%%%%%%%%%%%%%%%%%%%%%%%
	\section{Introduction}  
\begin{figure*}[htb]
	\centering
	\includegraphics[width=1.0\linewidth]{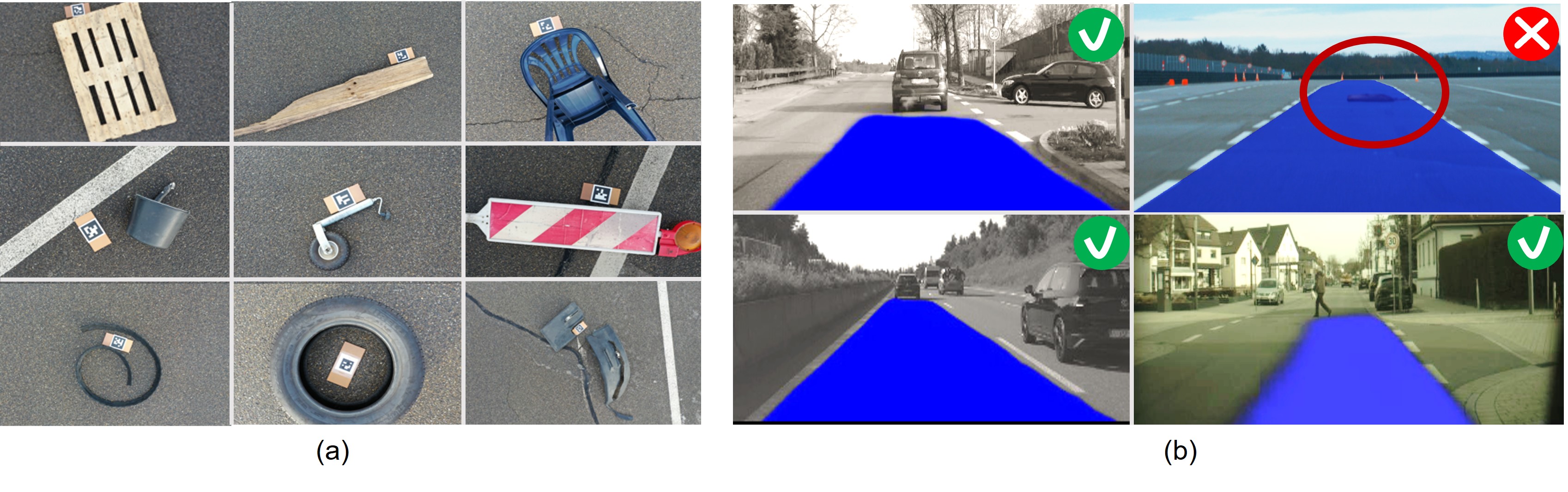}
        \vspace{-0.6cm}
	\caption{(a) Examples for unexpected obstacles found on highways, (b) Original AI corridor \cite{Michalke2021} that ends sharply at larger traffic participants. Application on different camera models. Top-right a tree stem is not detected by the original system.}
        \vspace{-0.4cm}
	\label{fig:LostCargo}
\end{figure*}

Detecting unexpected objects on roads is an essential task for automated vehicles to ensure the safety of car occupants. In the United States alone, it is estimated that between 2011 and 2014 more than 200,000 accidents were caused by road debris \cite{USRoadDebrisAccidents}. Similar for Germany, in 2021 in about 4500 cases (1.5\,\% of all accidents with personal injuries) an obstacle on the road was a root cause \cite{Bosch_AccidentResearch} (see Fig.~\ref{fig:LostCargo}a for a collection of obstacles lost on highways in Germany).

However, finding technical solutions for identifying these small objects has proven to be a difficult task. Radar systems, while good at detecting moving objects, are very limited in the detection of collision-relevant stationary obstacles. LiDAR systems, while great at detecting even small objects, cannot detect objects with low reflectance and objects at low height in far distances. On the other hand, AI-based video systems have been shown to achieve good results in detecting both stationary and dynamic objects even in severe weather (see e.g. \cite{Michalke2021}). However, even for video-based systems this task has proven to be a challenge. Traditional object detection methods focus on identifying objects of pre-defined classes. Their ability to detect unexpected objects of new, out-of-distribution classes is limited.

Recently, research has shifted  to the task of semantic segmentation to identify these outlier objects. Judging on a pixel basis, these methods try to distinguish between pixels belonging to one of the known object classes and those pixels belonging to outlier objects. These methods, however, still suffer from very poor performance both in terms of accuracy as well as real-time capability.

For this reason, we try to tackle this challenge with a different method: Indirect obstacle detection using a video-based AI ego-corridor segmentation \cite{Michalke2021}, which is a combined approach for lane and obstacle detection within the drivable corridor. The video-based AI ego-corridor provides a segmentation of the drivable area in the current lane. It offers a robust fallback for detecting ordinary traffic participants like cars and trucks by sharply cutting off the drivable area in front of them.

Our idea is to extend this approach in a way that also smaller unexpected obstacles are detectable based on the AI ego-corridor. This approach inverts the problem as we detect that the ego-corridor ends without being able to determine the object class or even a sharp bounding box around the object. Refer to Fig.~\ref{fig:LostCargo}b for the original AI corridor  \cite{Michalke2021} reacting to lane blockages caused by larger traffic participants and a tree stem. The approach allows an object-class-free detection of obstacles. It is able to detect larger objects even with out-of-distribution appearances as well. For smaller objects, detection is too late for a comfortable system reaction (see Fig. ~\ref{fig:LostCargo}b top-right image).

The remaining paper is structured as follows:
Chapter \ref{chap:related_work} discusses the related work and derives our contributions. Chapter \ref{chap:methods} gives a detailed description of our system approach. Two novel AI architectures are derived and different training methods are proposed to tackle the problem of restricted amount of data for ``lost obstacle'' on the road. In Chapter \ref{chap:Experiments}, we provide quantitative results for the detection quality and range when applied on previously unseen, smaller obstacles. Also, we evaluate the proposed architectures on our prototype vehicle.

	\section{Related Work}     \label{chap:related_work}
As in this contribution, we focus (a) on the detection of previously unseen, unexpected obstacles on the road and (b) pursue the idea of a combined lane and object detection, the following literature study will cover these two aspects.
\subsection{Unexpected Obstacle and Outlier Detection}
Most research in the field of unexpected obstacle detection has been done using depth information, as depth is a model-class-agnostic cue. Using stereo-camera images and Convolutional Neural Networks (CNNs), advances have been made to detect small obstacles by combining RGB and depth information to create a segmentation of the obstacles of interest. Depth from mono based on AI is no reliable option, as the key challenge is on out-of-distribution cases.  Measuring depth information directly requires the use of either a stereo camera or another source of depth information like a LiDAR sensor - both are expensive solutions with weaknesses in range.
%As a matter of completeness, depth from mono based on AI is no reliable option for out-of-distribution cases, as the present challenge of detecting unexpected obstacles.

Due to this, advances have been made to detect these obstacles without depth based on unexpensive sensors, as mono video RGB cameras. For the related area of outlier or anomaly detection, several approaches are known. Early efforts focused on uncertainty-based methods to estimate, which pixels constituted anomalies.

In \cite{Hendrycks2020ScalingOD, sml}, the authors used the statistics of the last layer (before softmax) of their CNN - the so-called logits. The CNN was trained to create a full semantic segmentation of traffic scenes. The authors worked under the hypothesis that pixels belonging to outliers statistically have lower logit values than their non-anomalous counterparts. While these methods are very simple to implement and computationally cheap, the outlier detection accuracy is very low.

The authors of \cite{fishyscapes} explored the idea of a ``void classifier'' that extends the task of semantic segmentation for $C$ predefined classes to $C+1$ classes where the new class represents a general anomaly class. They trained their network on the Cityscapes dataset \cite{Cordts2016} by changing the ground-truth of the pixels that had no ground-truth (pixels that did not fit into any of the pre-defined classes and were supposed to be ignored) to the new anomaly class.
Their method had very low anomaly detection accuracy, which showed that handling anomalies as a pre-defined class for semantic segmentation does not achieve robust performance.

In \cite{gan_synth}, the authors developed a reconstruction-based method for outlier detection. Feeding an RGB image depicting a traffic scene containing an unexpected obstacle to a generative adversarial network, the network creates a purposely wrong semantic segmentation of the scene over-segmenting the road (i.e., wrongly classifying obstacle pixels as road pixels). The GAN then uses this semantic segmentation to synthesize an image very similar to the original traffic scene, however crucially, missing the obstacles as the semantic segmentation had no information about these objects. The authors then use the difference between the original and the synthetic image to detect anomalous objects.
While this approach achieves state-of-the-art results, the computational requirements of the method render it unusable for real-time applications such as automated driving (more than 1 second inference time on Nvidia GTX 1080 Ti GPU).

In \cite{pebal}, the authors developed an outlier detection method based on abstention learning and energy-based models. Using the idea of a ``Gambler's loss'' from portfolio theory \cite{gamblersloss}, they train a semantic segmentation CNN to abstain from classifying outlier pixels as belonging to one of the pre-defined classes and to emit high ``energy'' for those outlier pixels. The energy scores of the pixels are then used to identify, whether a pixel is an anomalous one via simple thresholding.
Although this method, too, achieves state-of-the-art performance, the authors used a large CNN for their method that is not compatible with real-time requirements. Influence of downscaling on performance was not researched.

All these methods have in common, that outliers are treated and trained as an object class. In the following we present our alternative approach.
\subsection{AI Corridor: Combined Lane and Object Detection}
Traditionally, lane detection focusses on marked lanes. While different model-driven approaches have been developed that rely on hand-crafted features, in the last years deep learning (DL) approaches have achieved great performance improvements.

Deviating from a simple segmentation of lines or lane pixels, the authors of \cite{Michalke2021} developed a DL-based AI ego-corridor detection method that aims at understanding the current lane and traffic situation and segment only the currently drivable area of the ego vehicle's own lane. This work, which provides the basis of this contribution, compares a lane boundary detection algorithm to a corridor-based method in terms of robustness in situations, where lane boundary detection methods often face problems, such as roads without lane markings or bad environmental conditions (i.e., heavy rain).
The approach aims at implicitly detecting objects by detecting the end of the ego-corridor, which reduces the problem complexity, as the drivable corridor is a rather uniform object class. By utilizing the AI corridor and training it such that the model sets a sharp cut at its longitudinal edge in front of obstacles blocking the road, this and our approach do not aim at explicitly knowing the exact size and form of an obstacle. Simply put, the distribution of the well-defined class road corridor is used to help detecting out-of-distribution obstacles.

While their work shows that a DL ego-corridor approach can achieve good, robust performance for lane detection, we extend it in two major ways.
Firstly, their method uses very low-resolution gray-scale images consisting of less than 250.000 pixels and a very shallow CNN to achieve real-time performance in a test vehicle. In order to reach higher detection ranges for small objects, we used a more recent camera with high-resolution RGB images of more than 2 million pixels. Based on that, we rely on a more complex CNN architecture to accommodate the larger input and the information it carries. Naturally, increasing the input size by more than 10-fold increases the complexity required to process the information. However, we still aim for real-time performance on similar hardware. As such, we had to find a compromise between model complexity to accommodate the larger input size, and model inference time to keep within our real-time constraints.
Secondly, their ego-corridor approach is only capable of making good predictions for the longitudinal lane edge in ordinary traffic scenarios, i.e. when another car is in front or the road ends naturally. This, however, does not consider all possible situations that change the longitudinal edges of a lane - especially in inner city.

Therefore, our work aims at extending this capability to also make good predictions for the edge cases. Specifically, we aim at predicting the longitudinal edge of the drivable lane area in presence of unexpected, smaller obstacles such as ``lost cargo''. This extends the scope of the capability significantly, as our method does not need only to learn regular obstacles (i.e., cars, trucks, and motorcycles) that could cause the drivable area to change but instead needs to learn the very concept of obstacles that could come in all kinds of shapes and forms and block the road ahead. We hence extend the original architecture \cite{Michalke2021} to an ``AI corridor++'' to be able to react to smaller lost cargo obstacles and develop a second method that uses the AI corridor++ in combination with the work of \cite{pebal}. In this second approach, we combine explicit and implicit outlier detection in a combined network to create an AI corridor that is aware of unexpected obstacles (so-called ``multitask approach'', see Chapter \ref{chap:methods}).
\begin{table*}[ht!]
  \caption{Selected state-of-the-art approaches for the detection of unexpected obstacles.}
  \vspace{-0.2cm}
  \begin{center}
    \begin{tabularx}{1.0\textwidth}{lllXXXXXX}%{@{}lllllllr@{}}
      \toprule[1.25pt]
      Method   & Year & Ref.  & Architecture & Object-class-free Detection & Combined Usage of Lane Information & Detection Range / Quality for Unexpected Obstacles & Real-time capability \\ \midrule
      \'{H}endycks et al.  & 2020 & \cite{Hendrycks2020ScalingOD} & CNN, implicit SemSeg class  & No & No & Low / Low & Yes  \\
      Jung et al.  & 2021 & \cite{sml} & CNN, logits of SemSeg & No, implicit SemSeg class & No  & Low / Low & Yes  \\
      Blum et al.  & 2021 & \cite{fishyscapes} & CNN, SemSeg & No, explicit SemSeg class & No  & Low / Low & Yes  \\
      Blase et al.  & 2021 & \cite{gan_synth} & GAN, SemSeg & No, implicit model & No & Low / High & No   \\
      Yu et al.  & 2021 & \cite{pebal} & CNN, SemSeg, energy-based model & No, implicit model & No & Low / High & No   \\
      Michalke et al.  & 2021 & \cite{Michalke2021} & CNN & Partially (only large objects) & Yes & 50\,m / Low & Yes   \\
      \midrule
      \textbf{Ours: AI corridor++}  & & & \textbf{BiSeNet (context \& spatial path)} & \textbf{Yes} & \textbf{Yes} & \textbf{200\,m / High} & \textbf{Yes} \\
      \textbf{Ours: Multitask approach}  & & & \textbf{Two BiSeNets (shared context \& two spatial paths)} & \textbf{Yes} & \textbf{Yes} & \textbf{300\,m / High} & \textbf{Yes} \\
      \bottomrule[1.25pt]
    \end{tabularx}
  \end{center}
  \label{Tab:StateofRes}
   \vspace{-0.2cm}
\end{table*}
%online tested
%
Summarizing the known literature in the area of the detection of unexpected objects, Table \ref{Tab:StateofRes} shows that our work contributes the following novel aspects:
\begin{itemize}
	\item Object-class-free detection of obstacles: Problem inversion by evaluating the end of the ego-corridor,
	\item Combined usage of object and lane information,
	\item Low-cost sensor set: Mono video,
	\item High detection quality and range suitable for SAE L3 highway applications.
\end{itemize}
In the following section, we focus on two AI architectures that apply the idea of object-class-free obstacle detection.

	\section{Methods}             \label{chap:methods}
\begin{figure*}[]
	\centering
	\includegraphics[width=0.8\linewidth]{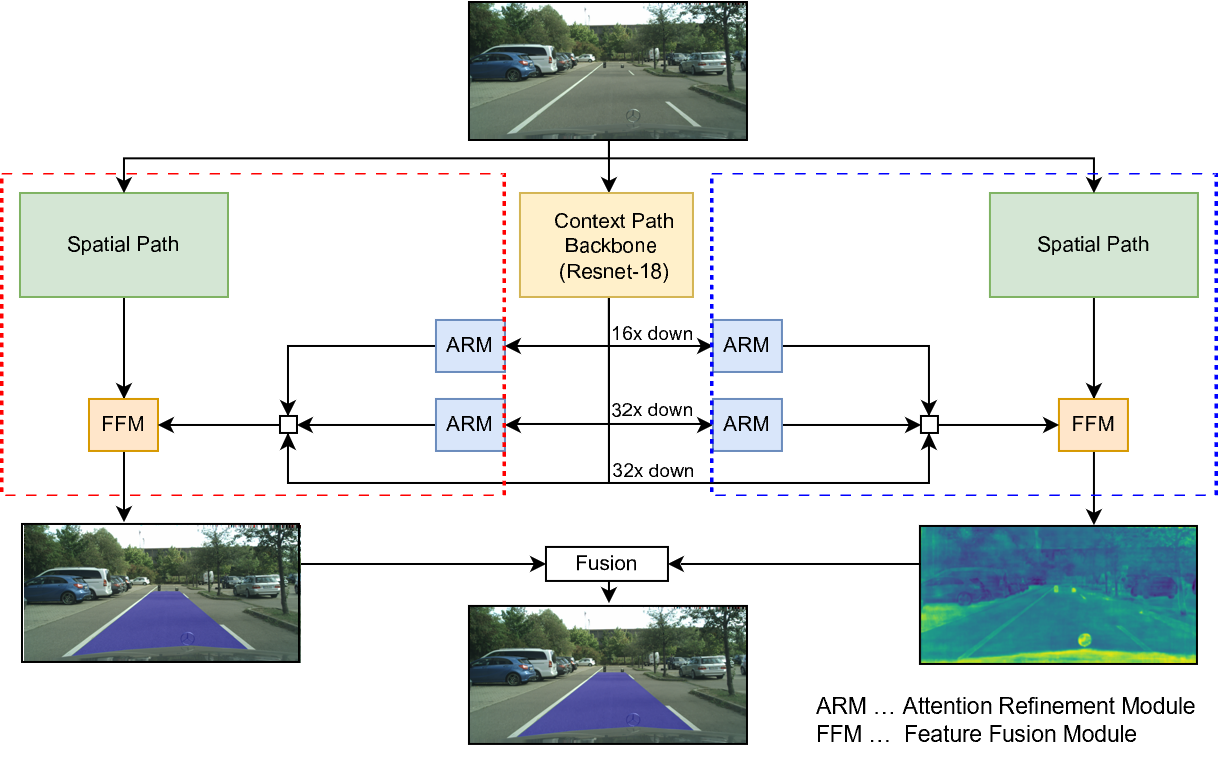}
	\caption{The BiSeNet-based network architecture of the multi-task approach. The Context Path backbone is shared between both paths. All modules in the red box are part of the ego-corridor segmentation, all modules in the blue box are part of the PEBAL-based outlier pixel segmentation. The ego-corridor-only approach (AI corridor++) utilizes the same architecture without the modules in the blue box. Test images taken from Daimler LostAndFound dataset \cite{LostAndFound2016}.}
        \vspace{-0.4cm}
	\label{fig:mtl_arch}
\end{figure*}
The following section describes the two approaches that were developed to extend the AI corridor to achieve unexpected obstacle detection capabilities. It further elaborates on the corridor representation and obstacle-aware data generation which is the most essential part of our approach.
\subsection{AI corridor++ CNN Architecture}
The original AI corridor solves the task of ego-corridor detection as a binary semantic segmentation task where each image pixel is classified either as part of the ego-corridor or as background. As the AI corridor++ merely aims to extend this approach with the focus on correct corridor segmentation in the presence of unexpected obstacles, the semantic segmentation approach remains unchanged. However, as the image input size has greatly increased, AI corridor++ requires an updated network architecture to account for the increased complexity. Further, as the goal is to run the method in an automotive system, the network architecture also has to achieve real-time performance.

To accommodate both of these priorities, we choose the BiSeNet \cite{bisenet} architecture with a Resnet-18 \cite{resnet} backbone for the ego-corridor segmentation. This dual-path semantic segmentation architecture consists of a deep path (Context Path) to extract low resolution, high-level contextual information and a shallow Spatial Path that extracts spatial information at a high resolution. It achieves near state-of-the-art performance on popular datasets like Cityscapes \cite{Cordts2016} while still achieving real-time performance on recent GPUs.

It is important to note, however, that the approach of AI corridor++ is not defined by the chosen CNN architecture but by the re-definition of the unexpected obstacle detection task as implicit detection with the ego-corridor and the data generation that results from this redefinition. As such, any modern semantic segmentation network is suitable.
\subsection{Multitask Approach}
Safety-critical applications like automated driving greatly benefit and may even require redundancy. For this purpose and to evaluate synergy effects, we also developed a second approach for the AI Corridor++ that combines the definition of the problem as implicit detection with a more traditional approach of explicit obstacle detection.

\begin{figure}[]
	\centering
	\begin{subfigure}{.23\textwidth}
		\centering
		\setlength{\fboxsep}{0pt}\fbox{\includegraphics[width=.99\linewidth]{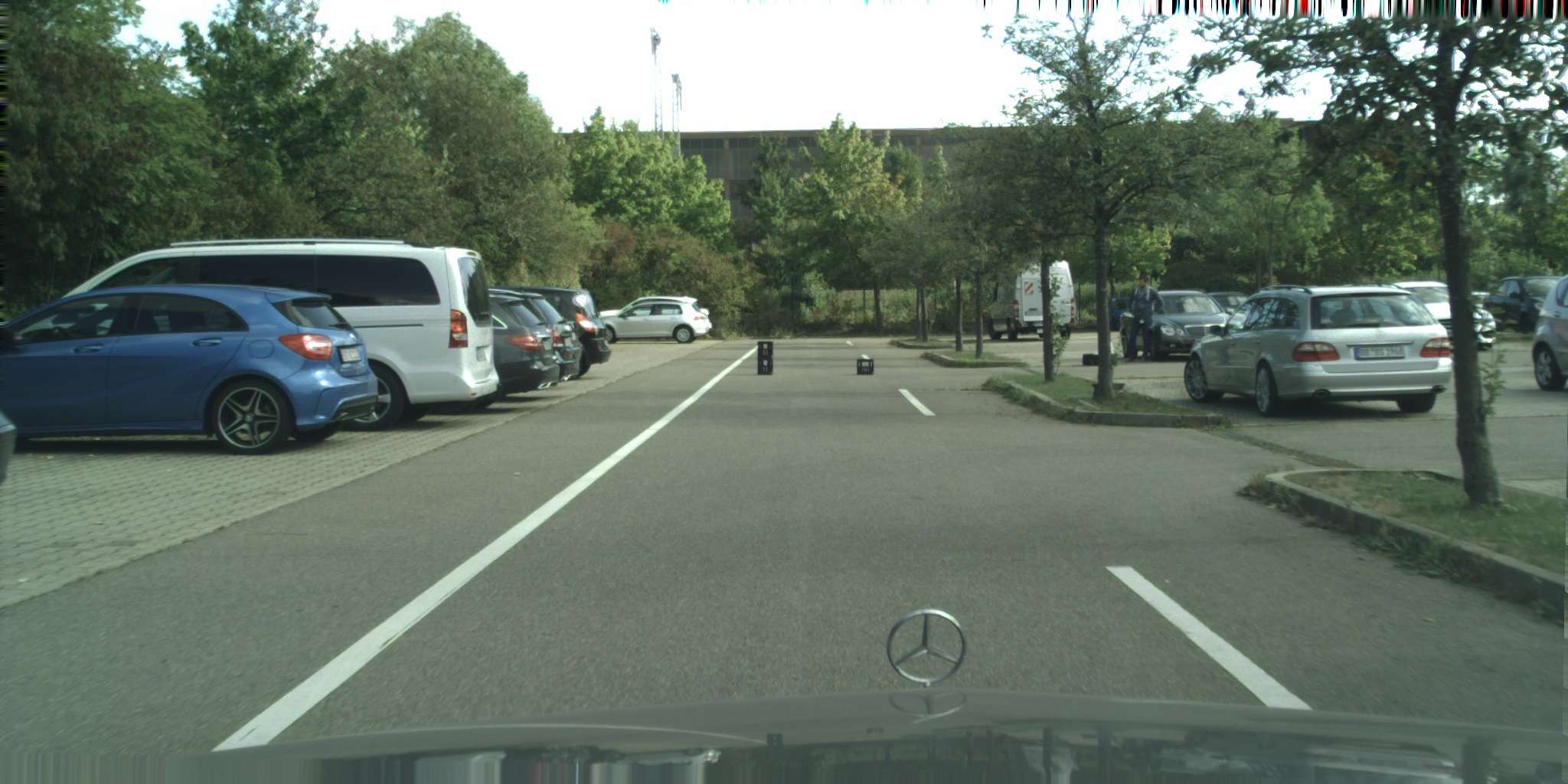}}
		\caption{Input image}
		\label{fig:orig_image_pebal}
	\end{subfigure}
	\begin{subfigure}{.23\textwidth}
		\centering
		\setlength{\fboxsep}{0pt}\fbox{\includegraphics[width=.99\linewidth]{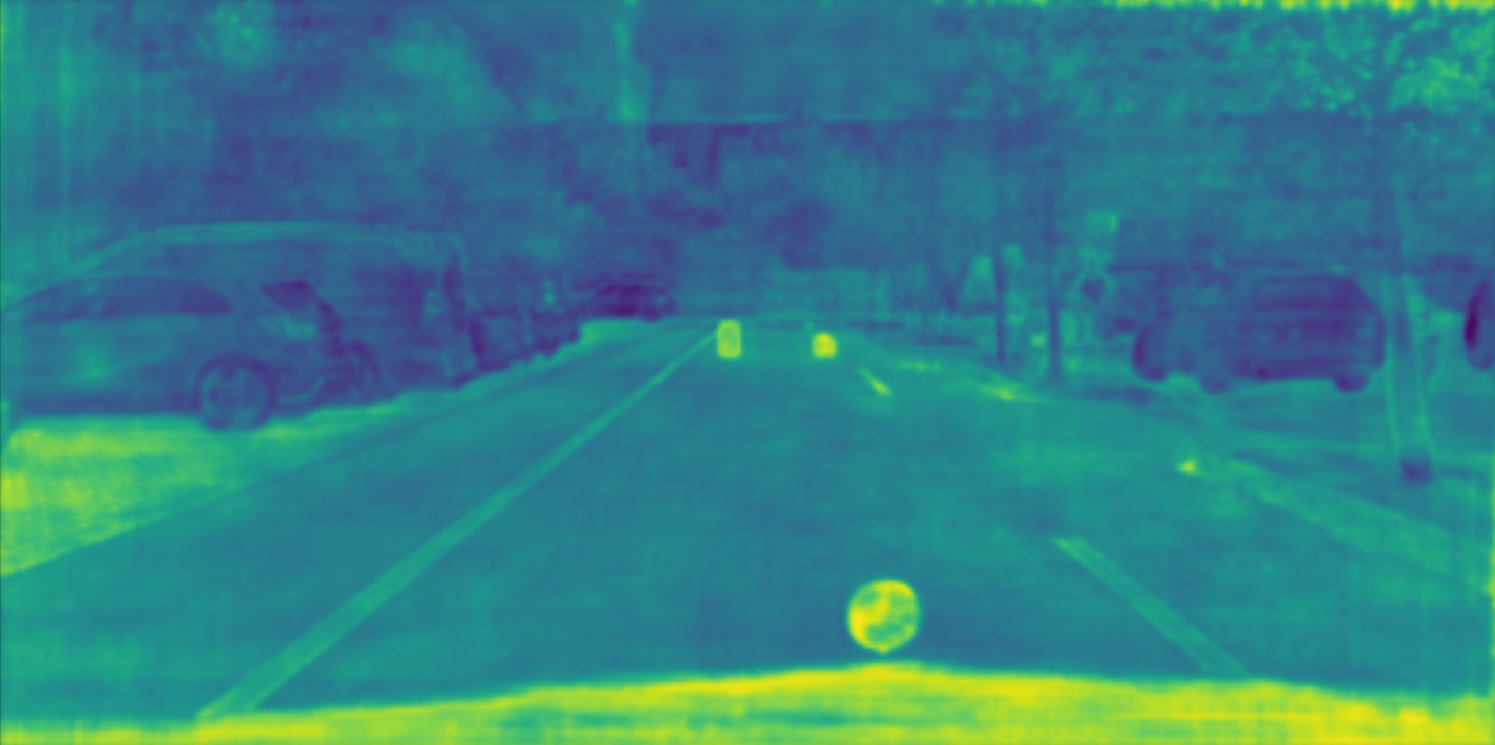}}
		\caption{Energy Map}
		\label{fig:energy_map_pebal}
	\end{subfigure}
	\caption{Illustration of energy map output of PEBAL. Outliers are found based on thresholding.}
        \vspace{-0.4cm}
	\label{fig:pebal}
\end{figure}

The approach we chose to fill this supplementary role is based on Pixel-wise Energy-biased Abstention Learning (PEBAL) \cite{pebal}. We scaled it down to tit our real-time application. PEBAL aims at teaching a full scene semantic segmentation network to abstain from classifying outlier pixels as one of the pre-defined inlier classes of a full scene semantic segmentation task. It further teaches the network to learn a high disparity of energy (general uncertainty of belonging to any inlier class) between inlier and outlier pixels. The output of this method is pixelwise classification of inlier and outlier pixels based on a chosen energy threshold (see Fig. \ref{fig:pebal}).
For details regarding PEBAL, please refer to \cite{pebal}.
As recent research\cite{yolop,hybridnets} has shown that multi-task networks can increase the performance of the individual tasks, we implement the two approaches by extending the BiSeNet architecture with another path for the PEBAL-based explicit unexpected obstacle detection. This has the additional benefit of drastically reducing the runtime compared to running these two tasks in two separate networks.

The two outputs of this multi-task network - ego-corridor mask and explicit unexpected obstacle pixels - are then fused to refine the ego-corridor if the object was missed by the ego-corridor and intersection between corridor mask and outlier pixels exists (see Fig. \ref{fig:mtl_arch}).
\subsection{AI Corridor++ Data Generation}
The most essential part of our approach is the representation and data generation of the corridor. Like in the original work \cite{Michalke2021}, the corridor is defined as the drivable space within the ego-lane. Unlike the original work, the longitudinal reach of this drivable space does no longer only aim to go up to the first traffic participant (i.e., car, motorcycle, truck) but instead sets a focus to extend this longitudinal reach to end in front of any obstacle that would hinder a secure driving on the road.

\begin{figure}[]
	\centering
	\begin{subfigure}{.15\textwidth}
		\centering
		\setlength{\fboxsep}{0pt}\fbox{\includegraphics[width=.99\linewidth]{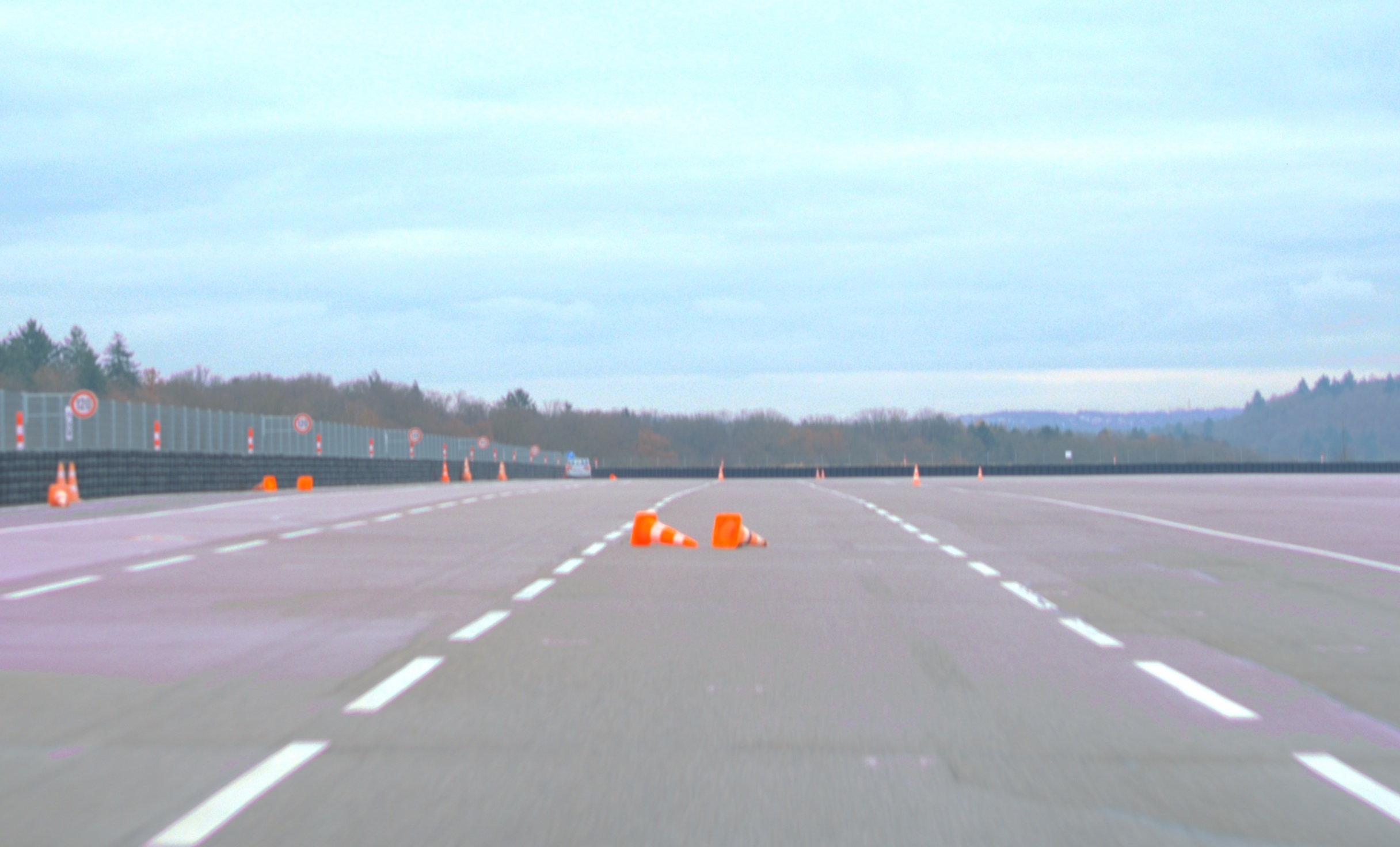}}
		\caption{Input image}
		\label{fig:input_format}
	\end{subfigure}
	\begin{subfigure}{.15\textwidth}
		\centering
		\setlength{\fboxsep}{0pt}\fbox{\includegraphics[width=.99\linewidth]{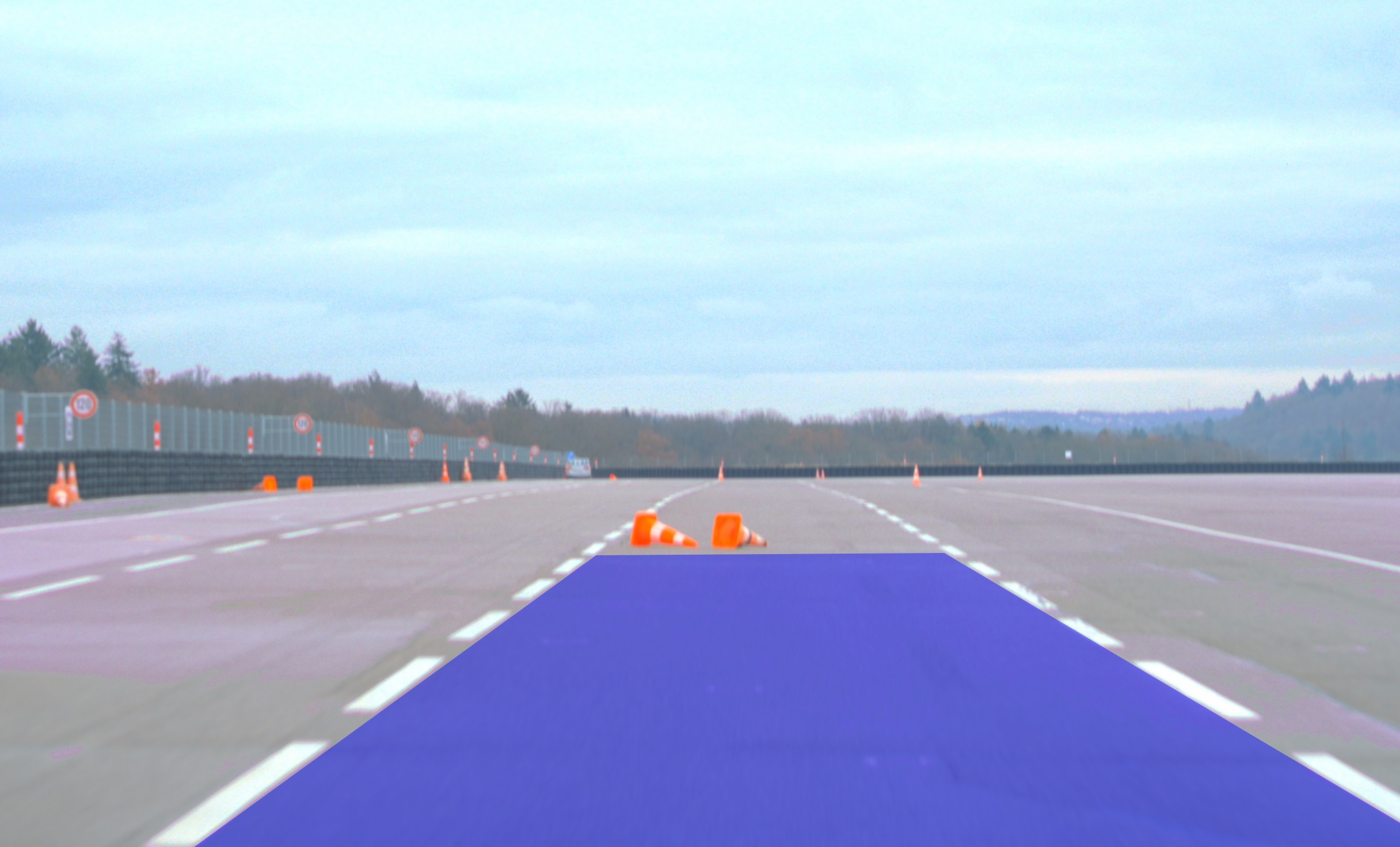}}
		\caption{Overlay output}
		\label{fig:overlay_format}
	\end{subfigure}
	\begin{subfigure}{.15\textwidth}
		\centering
		\setlength{\fboxsep}{0pt}\fbox{\includegraphics[width=.99\linewidth]{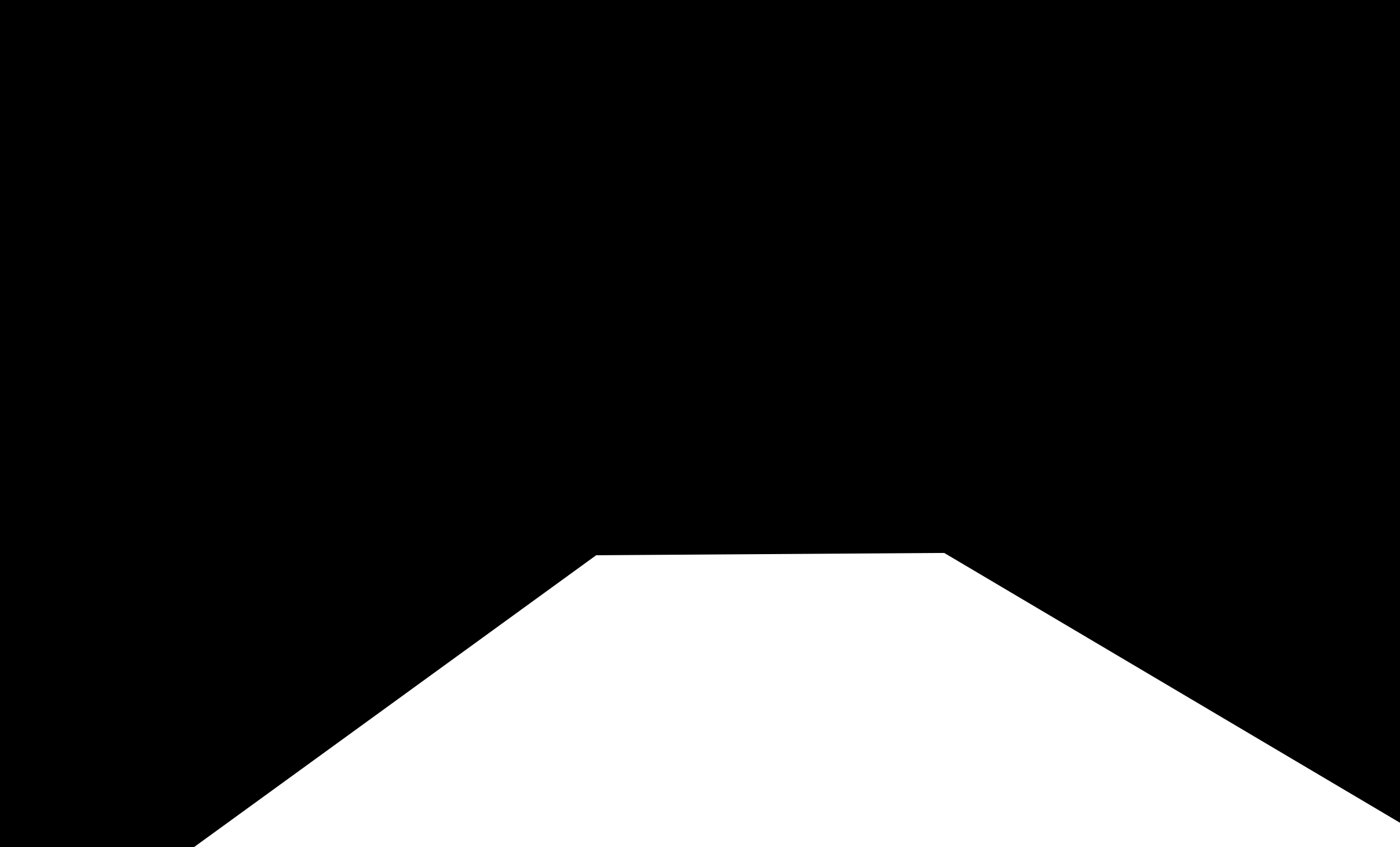}}
		\caption{Mask output}
		\label{fig:output_format}
	\end{subfigure}
	\caption{Image input and ego-corridor data format.}
        \vspace{-0.6cm}
	\label{fig:data_format}
\end{figure}

To achieve this capability extension, we created a comprehensive dataset that contains 1) ordinary traffic scenes to ensure good performance in generic traffic scenarios and 2) traffic scenes with real and synthetically generated obstacles found or placed on roads.
Figure \ref{fig:data_format} illustrates such a traffic scene and our generated corridor data representation.
\subsection{Synthetic Data Generation}
While we created a substantial dataset with many objects with very different characteristics, the creation of a large-scale dataset with full variety of scenarios and obstacles is both cumbersome and time-consuming. A large dataset with a high variety of obstacles, however, is necessary to capture the practically infinite variety of obstacles that one might encounter on the road. To increase our variety of obstacles and to supplement the traffic scenes with real obstacles, we choose a method of synthetic data generation.

We created a dataset of synthetic objects by using the labelled mask of the objects found in the images of the COCO dataset \cite{coco}, a large object detection and segmentation dataset consisting of more than 200.000 images with ground-truth masks for objects of 80 different classes. As 20 classes are traffic-relevant, objects of 60 classes remained for us to use for the generation of traffic scenes with synthetic out-of-distribution objects.

\begin{figure}[]
	\centering
	\includegraphics[width=0.99\linewidth]{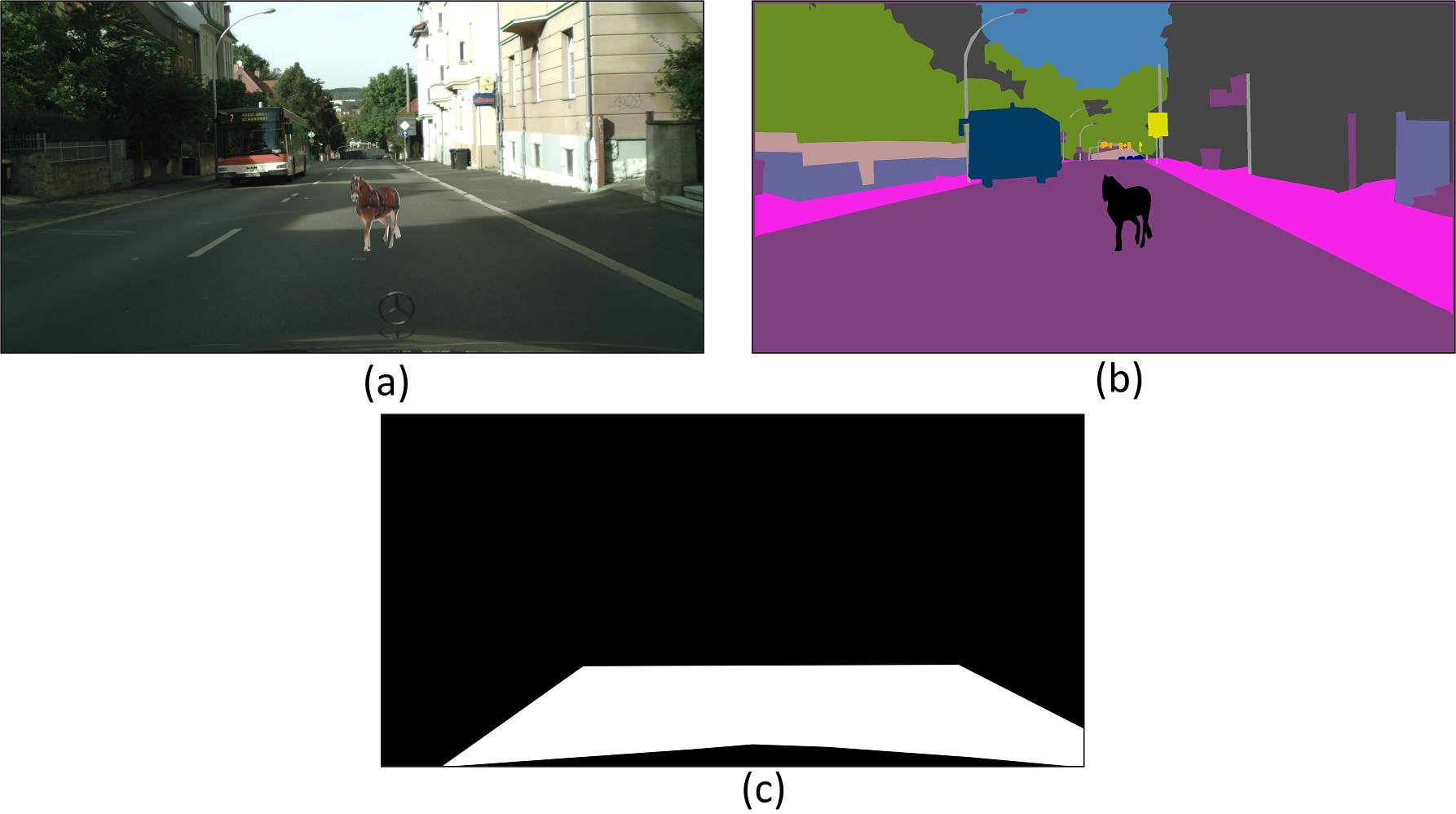}
	\caption{Traffic scene with a synthetic object (a). As the object changes the scene, the
		ground-truth labels for full scene semseg (b) and ego-lane
		segmentation (c) are adjusted as well.}
        \vspace{-0.2cm}
	\label{fig:example_synthetic}
\end{figure}

During the training process with our original traffic scenes both from our own dataset with real obstacles as well as from a dataset with ordinary traffic scenes without any obstacles, these synthetic obstacles were extracted from their original image using their provided masks and then inserted into the images, changing both the image as well as the ground-truth labels used for training. To ensure a smooth transition of the objects into the traffic scenes, we used the augmentation techniques described in \cite{fishyscapes}. As position, size, rotation, etc. of these obstacles are chosen randomly (with conditions to ensure traffic relevancy), we have a nearly infinite amount of different traffic scenes with synthetic obstacles, greatly increasing the potential object representation in the dataset. Figure \ref{fig:example_synthetic} illustrates exemplarily the insertion of a synthetic foal into a traffic scene as well as the resulting changes to the ground-truth labels.
\subsection{Corridor Post-Processing}
\begin{figure}[h!]
	\centering
	\begin{subfigure}{.23\textwidth}
		\centering
		\setlength{\fboxsep}{0pt}\fbox{\includegraphics[width=.99\linewidth]{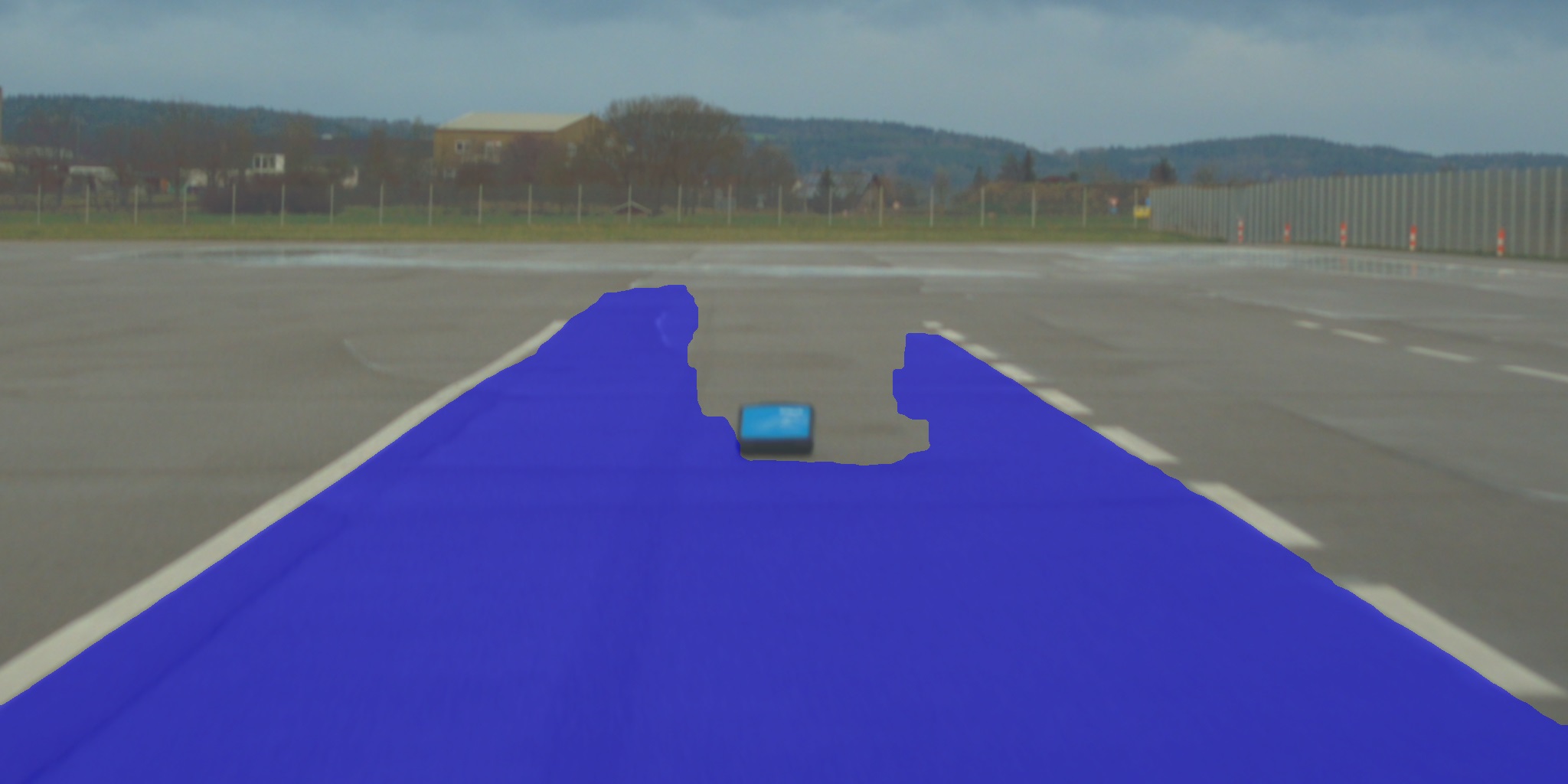}}
		\caption{No Post-Processing}
		\label{fig:bad}
	\end{subfigure}
	\begin{subfigure}{.23\textwidth}
		\centering
		\setlength{\fboxsep}{0pt}\fbox{\includegraphics[width=.99\linewidth]{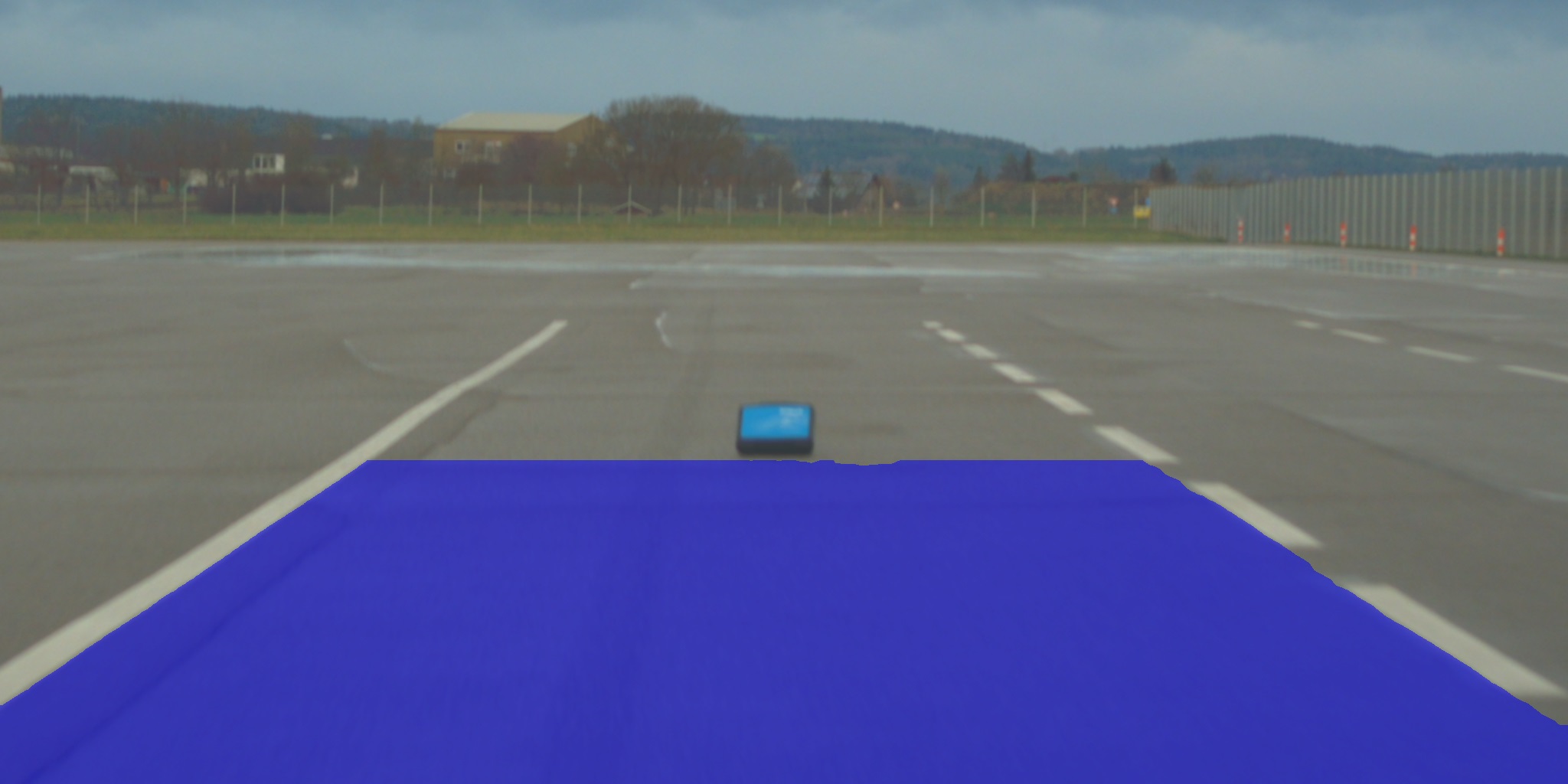}}
		\caption{After Post-Processing}
		\label{fig:good}
	\end{subfigure}
	\caption{Example of the ego-corridor obstacle detection
		without correct ego-lane limitation and its correction with post-processing.}
        \vspace{-0.4cm}
	\label{fig:post_processing}
\end{figure}

Ideally, the CNN model directly outputs an ego-lane mask that by itself is a good approximation of the underlying ground-truth and requires no post-processing. To some extend, however, the network outputs an ego-corridor mask that may have small openings and that hence is not one contiguous area. Especially with obstacles close to the ego vehicle, the AI corridor correctly detects the obstacle but without the correctly limiting of the ego-corridor over the full lane width. Figure \ref{fig:post_processing} shows an example of such a case. To overcome these problems, we apply simple contour finding as well as a search for a sudden drop of width to find a continuous corridor and smoothen the longitudinal edge of the corridor. More sophisticated post-processing approaches are conceivable and could improve performance further. Based on the sharp edge, the following system modules can create an object candidate that is handle in a fallback path. This path complements the performance path for object detection in our automated driving stack (for details on how the fallback path is coupled to the system refer to \cite{Michalke2021}).
\begin{table*}[]
\caption{Ego-lane segmentation results (detection rates) for the trained models for objects at 25\,m, 50\,m, 100\,m, 200\,m, and 300\,m distance. For each of the distances 3 images of each of the 28 ``lost cargo'' test items were used. For proof of robustness against false-positive detections, 12 runs on the test track without lost cargo objects were evaluated with roughly 200 images for each run. The trained object detector created only few uncritical false-positive detections in larger distances that could be alleviated by object tracking.}
\begin{center}
	\begin{tabular}{lrrrrr} \toprule
		Method & 25\,m        & 50\,m          & 100\,m         & 200\,m         & 300\,m   \\ \midrule
		\textit{AI corridor++}&&&&&\\
		Naive     & 11 (13.1\,\%)& 48 (57.1\,\%)& 55 (65.4\,\%)& \textbf{76 (90.4\,\%)}&76 (90.4\,\%) \\
		Obstacle  &   36 (42.8\,\%) &\textbf{ 61 (72.0\,\%)} & \textbf{56 (66.6\,\%)}&\textbf{76 (90.4\,\%)} & \textbf{80 (95.2\,\%)}    \\
		Synthetic &     21 (25.0\,\%)      & 42 (50.0\,\%)  &  38 (45.2\,\%)    & 57 (67.8\,\%) &68 (80.9\,\%)\\ \midrule
		\textit{AI corridor++ \& PEBAL fusion}             & \textbf{48 (57.1\,\%)} & 34 (40.4\,\%) & 6 (7.1\,\%) & 22 (26.1\,\%) & 35 (41.6\,\%) \\ \bottomrule
	\end{tabular}
  \end{center}
	\label{table:lostcargo_results}
       \vspace{-0.5cm}
\end{table*}

	\section{Experiments}      \label{chap:Experiments}
\begin{figure*}
	\centering
	\begin{subfigure}{0.8\textwidth}
		\centering
		\setlength{\fboxsep}{0pt}\fbox{\includegraphics[width=.99\linewidth]{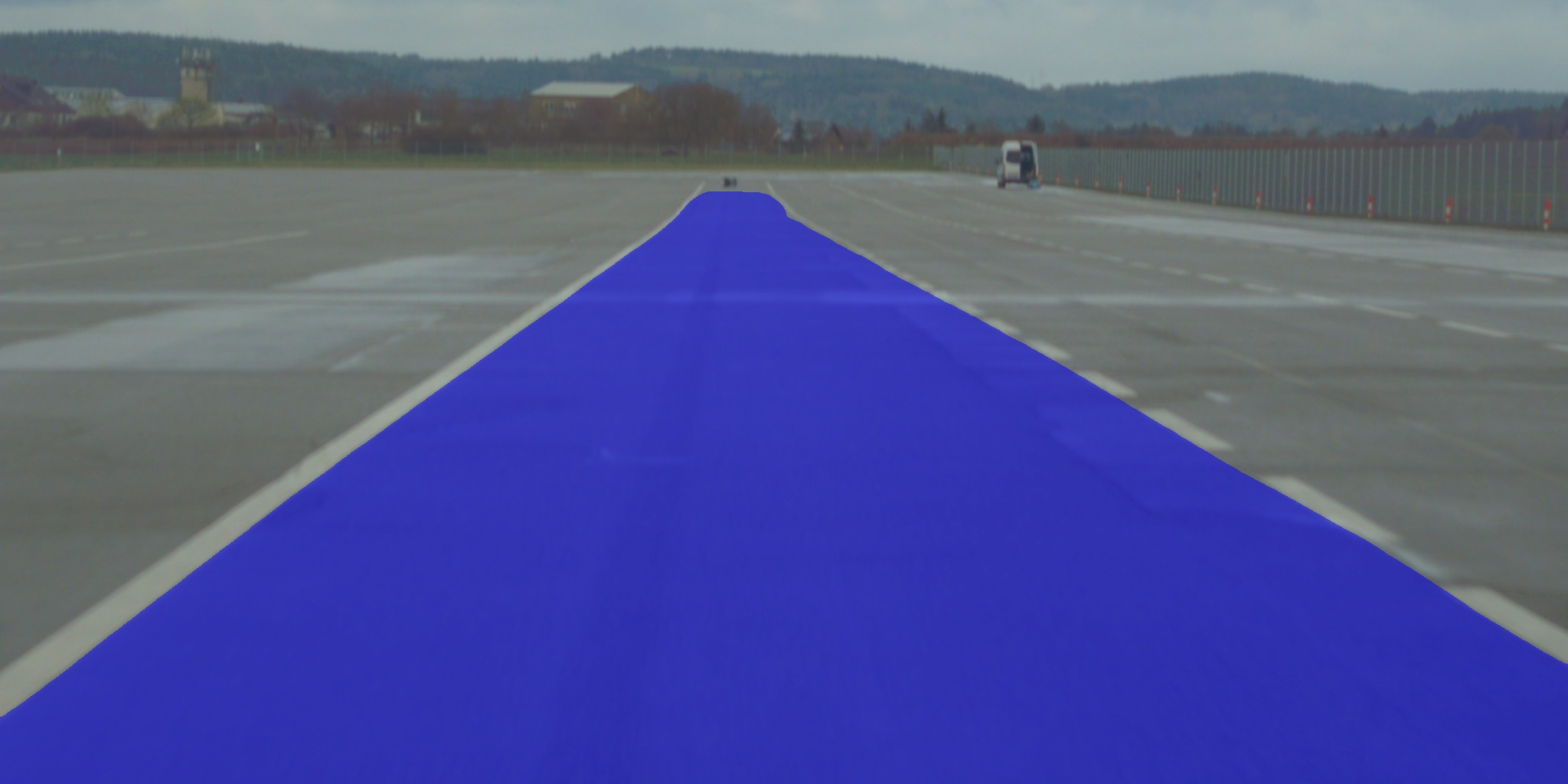}}
		\caption{Correct ego-lane segmentation for object at 300\,m distance}
		\label{fig:pic1}
	\end{subfigure}
	\begin{subfigure}{.20\textwidth}
		\centering
		\setlength{\fboxsep}{0pt}\fbox{\includegraphics[width=.99\linewidth]{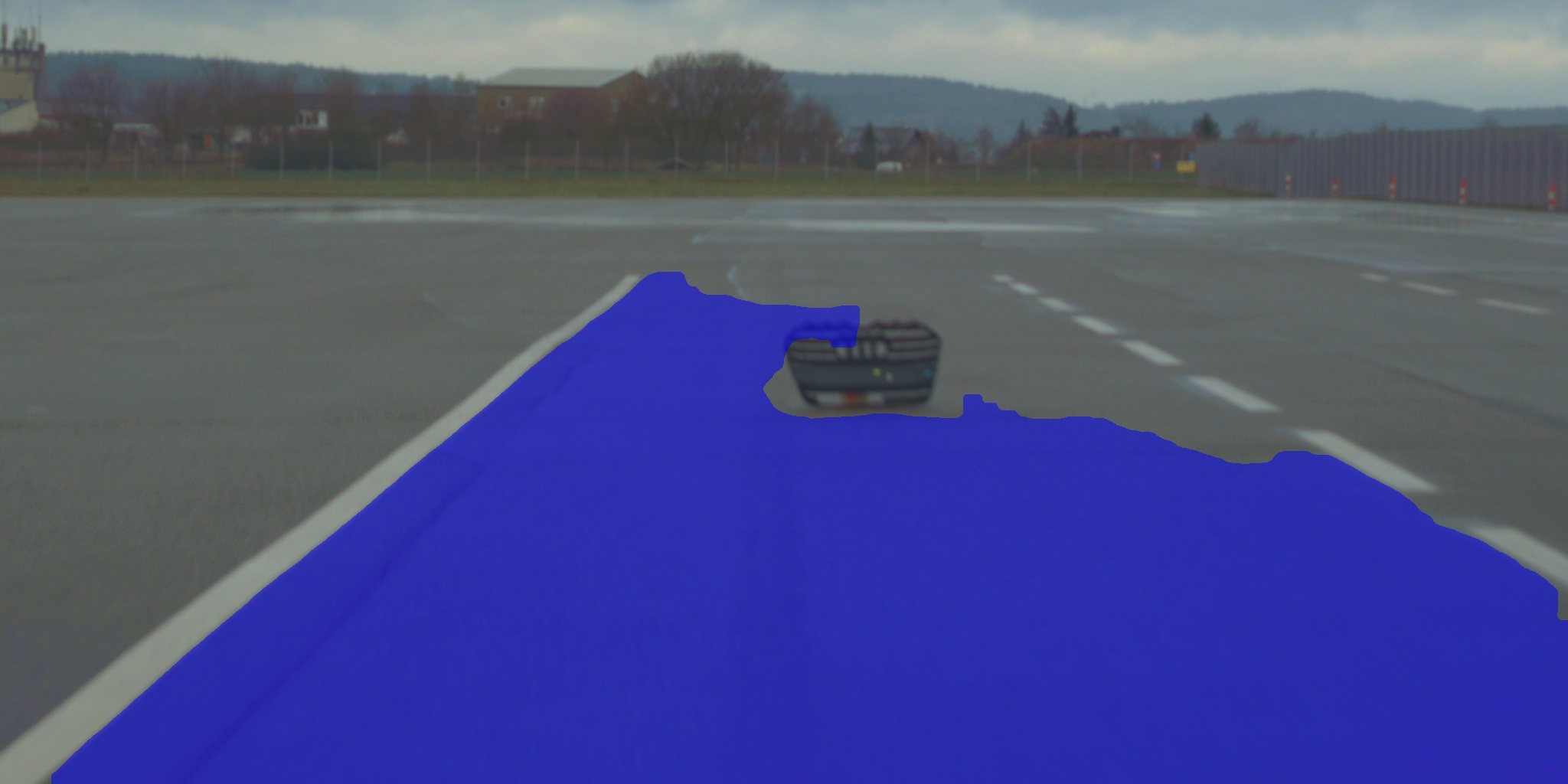}}
		\caption{No PP}
		\label{fig:ok1}
	\end{subfigure}
	\begin{subfigure}{.20\textwidth}
		\centering
		\setlength{\fboxsep}{0pt}\fbox{\includegraphics[width=.99\linewidth]{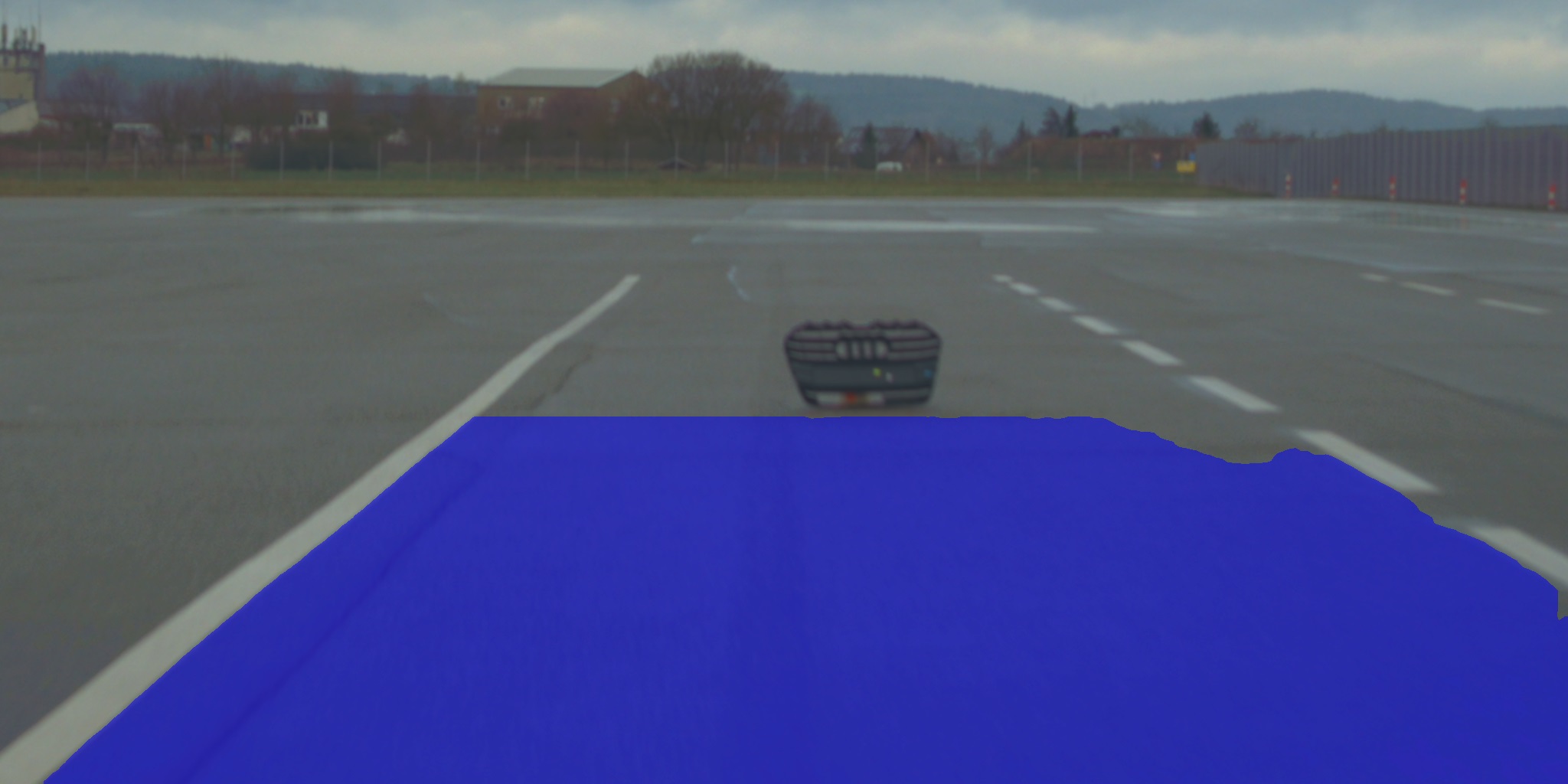}}
		\caption{With PP}
		\label{fig:ok2}
	\end{subfigure}
	\begin{subfigure}{.20\textwidth}
		\centering
		\setlength{\fboxsep}{0pt}\fbox{\includegraphics[width=.99\linewidth]{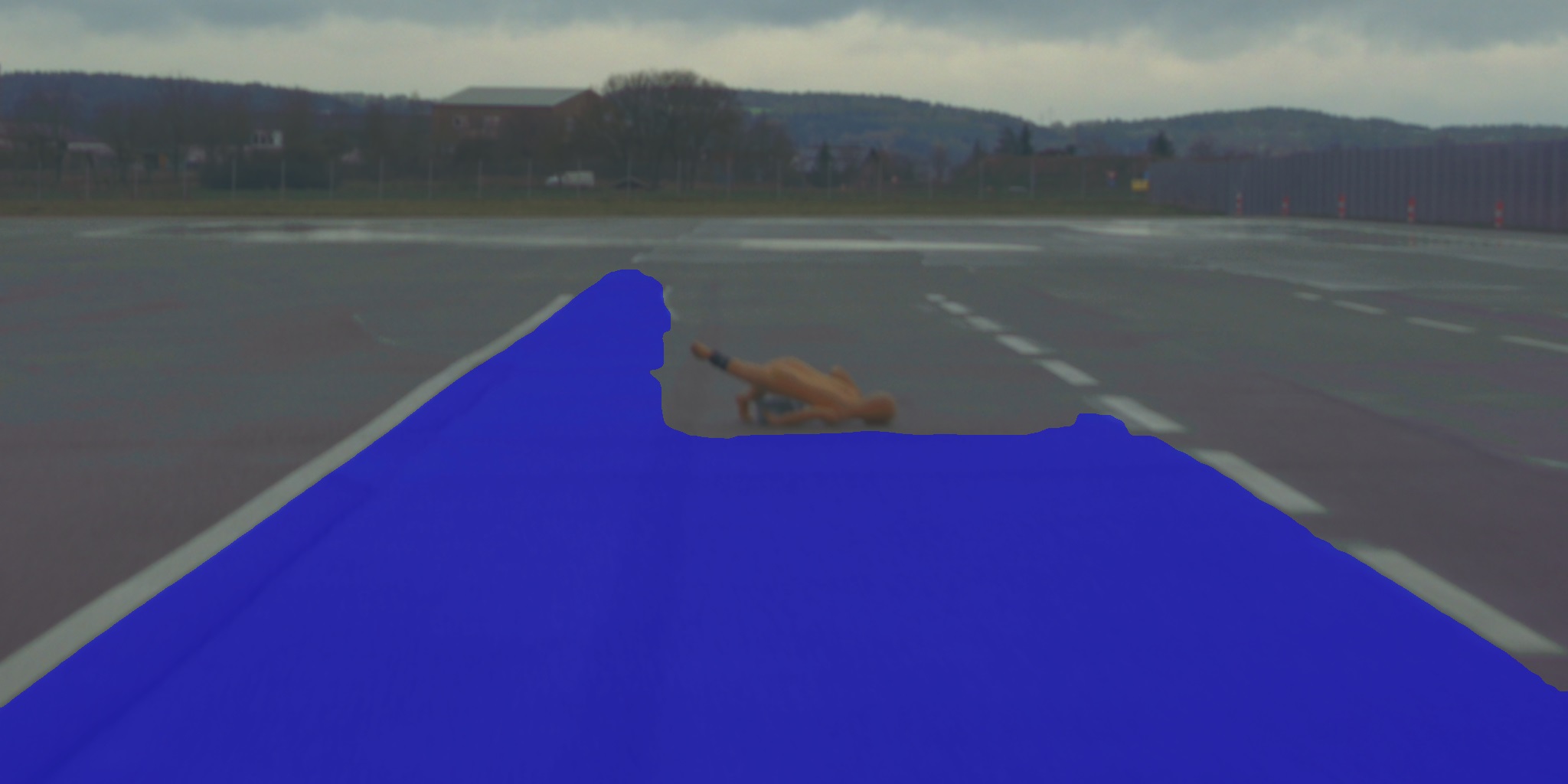}}
		\caption{No PP}
		\label{fig:ok3}
	\end{subfigure}
	\begin{subfigure}{.20\textwidth}
		\centering
		\setlength{\fboxsep}{0pt}\fbox{\includegraphics[width=.99\linewidth]{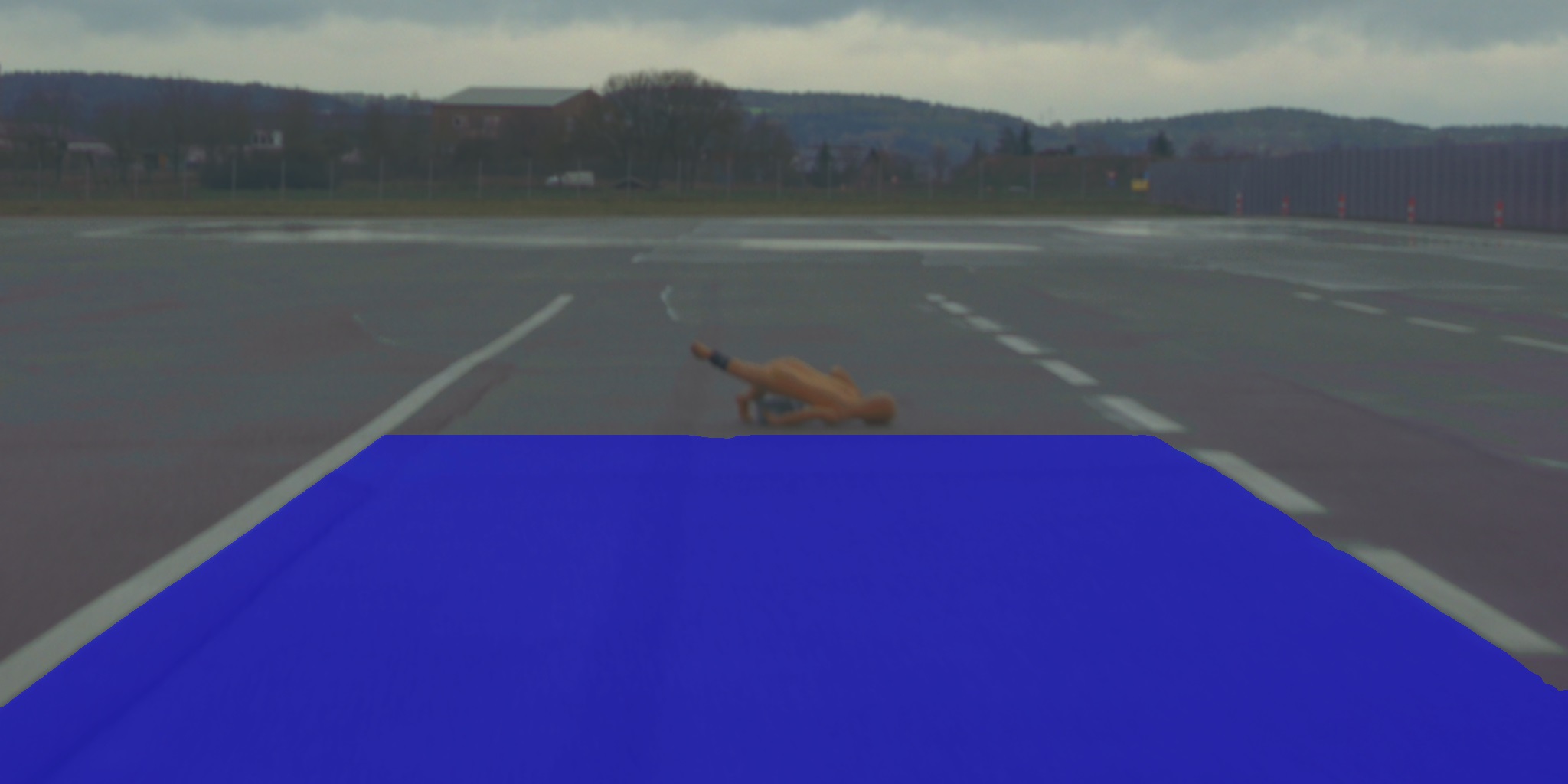}}
		\caption{With PP}
		\label{fig:ok4}
	\end{subfigure}
	\caption{Example ego-corridor segmentation for an object at 300\,m distance and illustration of post-processing (PP) effects.}
        \vspace{-0.4cm}
	\label{fig:obst_qualitative_res}
\end{figure*}
In the following section, we discuss the experiments conducted to evaluate our two approaches and present qualitative as well as quantitative results.
\subsection{Datasets}
For the training of the ego-corridor, we conduct experiments on a dataset consisting of three parts:
\begin{enumerate}
	\item Self-recorded ordinary traffic scenes captured by a tele mono-video camera consisting of 7000 labelled images
	\item Self-recorded traffic scenes with out-of-distribution obstacles captured by the same tele mono-video camera consisting of 3500 labelled images
	\item A subset of the BDD100k dataset \cite{Yu2018}, a popular traffic scene dataset with labels for a multitude of video perception tasks including drivable space segmentation. We use the images that have a drivable space label to supplement our dataset for better generalization performance. This part consists of 20.000 images and only the labels for the ego-drivable space were used and post-processed to match our ground-truth semantics.
\end{enumerate}
As we had no full scene semantic segmentation labels available for our datasets at the time of this work, we used Cityscapes dataset \cite{Cordts2016} for the training of the PEBAL path in conjunction with synthetic obstacles extracted from the COCO dataset.
\subsection{Implementation Details}
The networks were implemented in Tensorflow 2 and run as a Robot Operating System (ROS) node as part of a complete automated driving software stack on a test vehicle.
We train the ego-corridor semantic segmentation using the Focal Loss. The PEBAL path is trained using the hyperparameters described in \cite{pebal}. For the multitask network, we train both paths independently from each other, first training the network on the ego-corridor segmentation and then training the PEBAL path with a frozen backbone.
\subsection{Experimental Setup}
\begin{figure}[]
	\centering
	\begin{subfigure}{.22\textwidth}
		\centering
		\setlength{\fboxsep}{0pt}\fbox{\includegraphics[width=.99\linewidth]{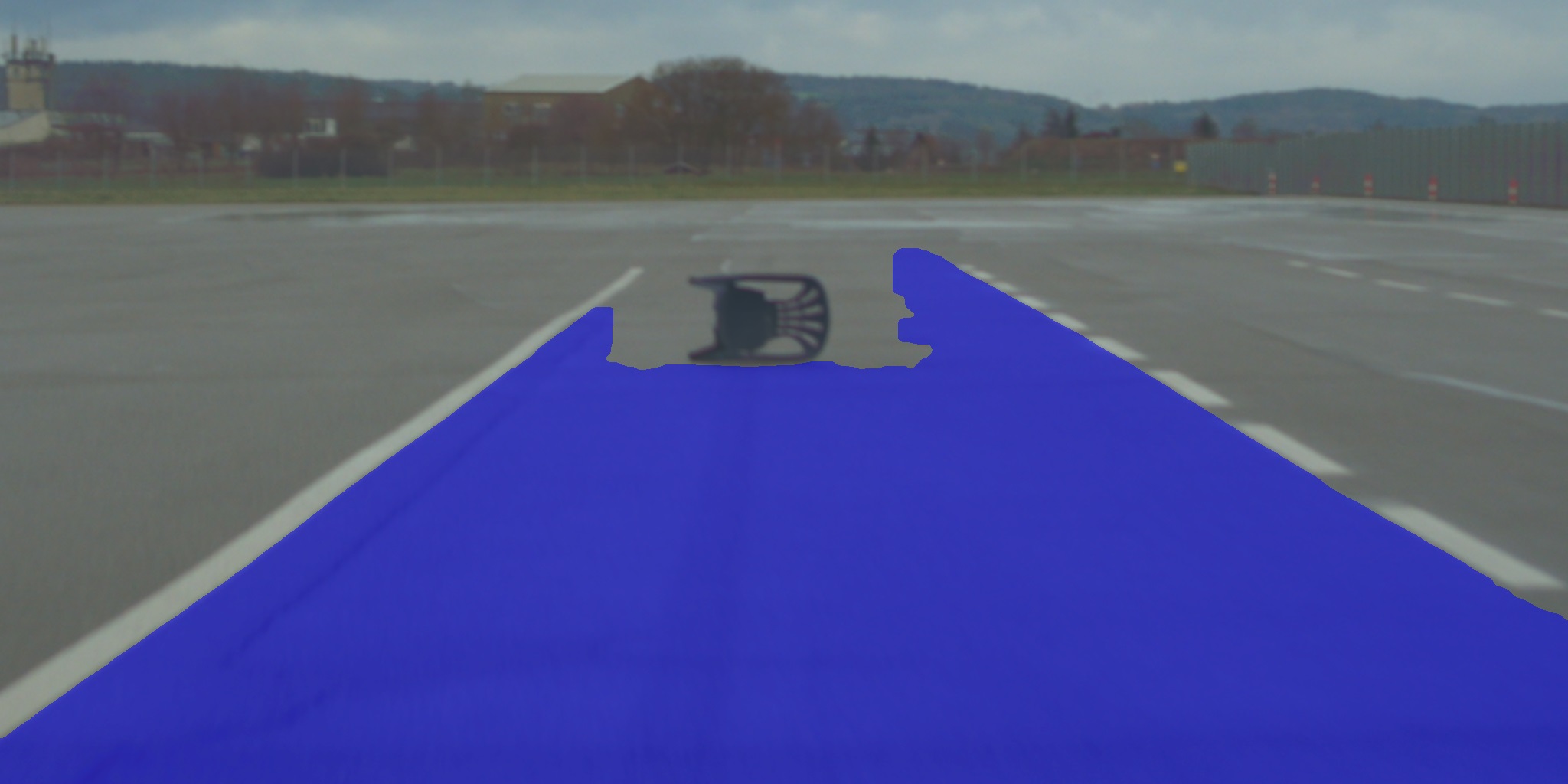}}
		\caption{Incorrect Detection}
		\label{fig:incorrect_det}
	\end{subfigure}
	\begin{subfigure}{.22\textwidth}
		\centering
		\setlength{\fboxsep}{0pt}\fbox{\includegraphics[width=.99\linewidth]{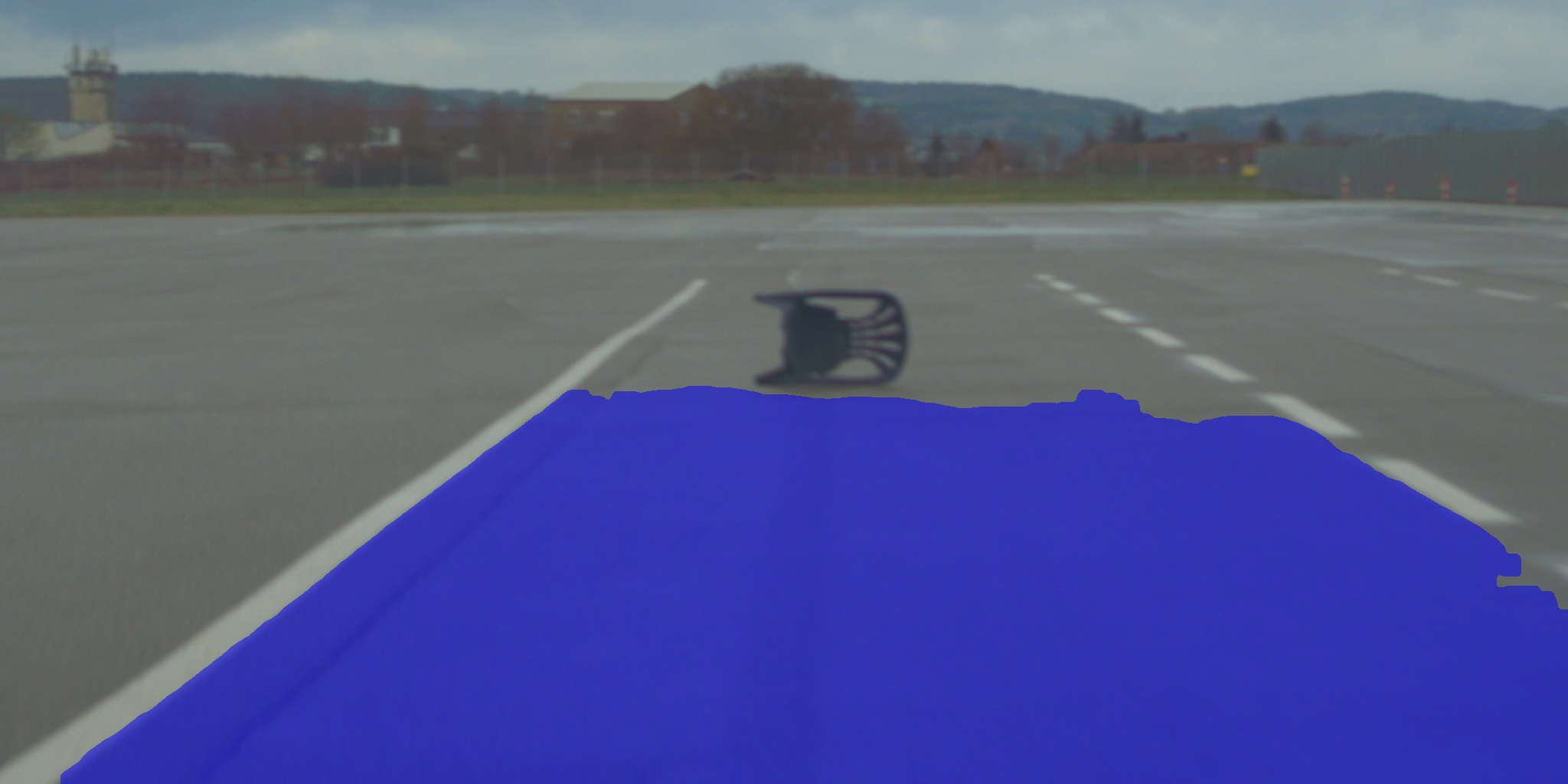}}
		\caption{Correct Detection}
		\label{fig:correct_det}
	\end{subfigure}
	\caption{Examples for correct (right) and incorrect (left) ego-corridor segmentation. As the ego-corridor merely wraps around the object and continues far beyond the object, we consider the left example to be unsuccessful.}
        \vspace{-0.6cm}
	\label{fig:detection_examples}
\end{figure}

To evaluate our two approaches, we create a test set of scenarios on a test track involving objects that were not seen during the training process. These 28 different objects varied considerably in size, shape and color.
We record one drive per object and evaluate the models' detection performance at five different distances between the car and the objects: 25\,m, 50\,m, 100\,m,
200\,m, and 300\,m. We measure whether the longitudinal edge of the corridor is correctly set (with a fixed amount of maximum deviation allowed) in front of the object so that an under-segmentation of the corridor also is penalized. Figure \ref{fig:detection_examples} illustrates what constitutes a correct and an incorrect detection.

We compare 4 different variants of our approaches:
\begin{enumerate}
	\item The ego-corridor-only approach trained a) with only ordinary traffic scenes (\textit{Naive}), b) with traffic scenes containing real obstacles (\textit{Obstacle}) and c) synthetic objects inserted into traffic scenes (\textit{Synthetic})
	\item The multitask approach using the ego-corridor trained with real obstacles in combination with the PEBAL-based explicit outlier detection.
\end{enumerate}
\begin{figure*}
	\centering
	\begin{subfigure}{.2\textwidth}
		\centering
		\small\textbf{Ego-Corridor Path}\par\medskip
		\setlength{\fboxsep}{0pt}\fbox{\includegraphics[width=.99\linewidth]{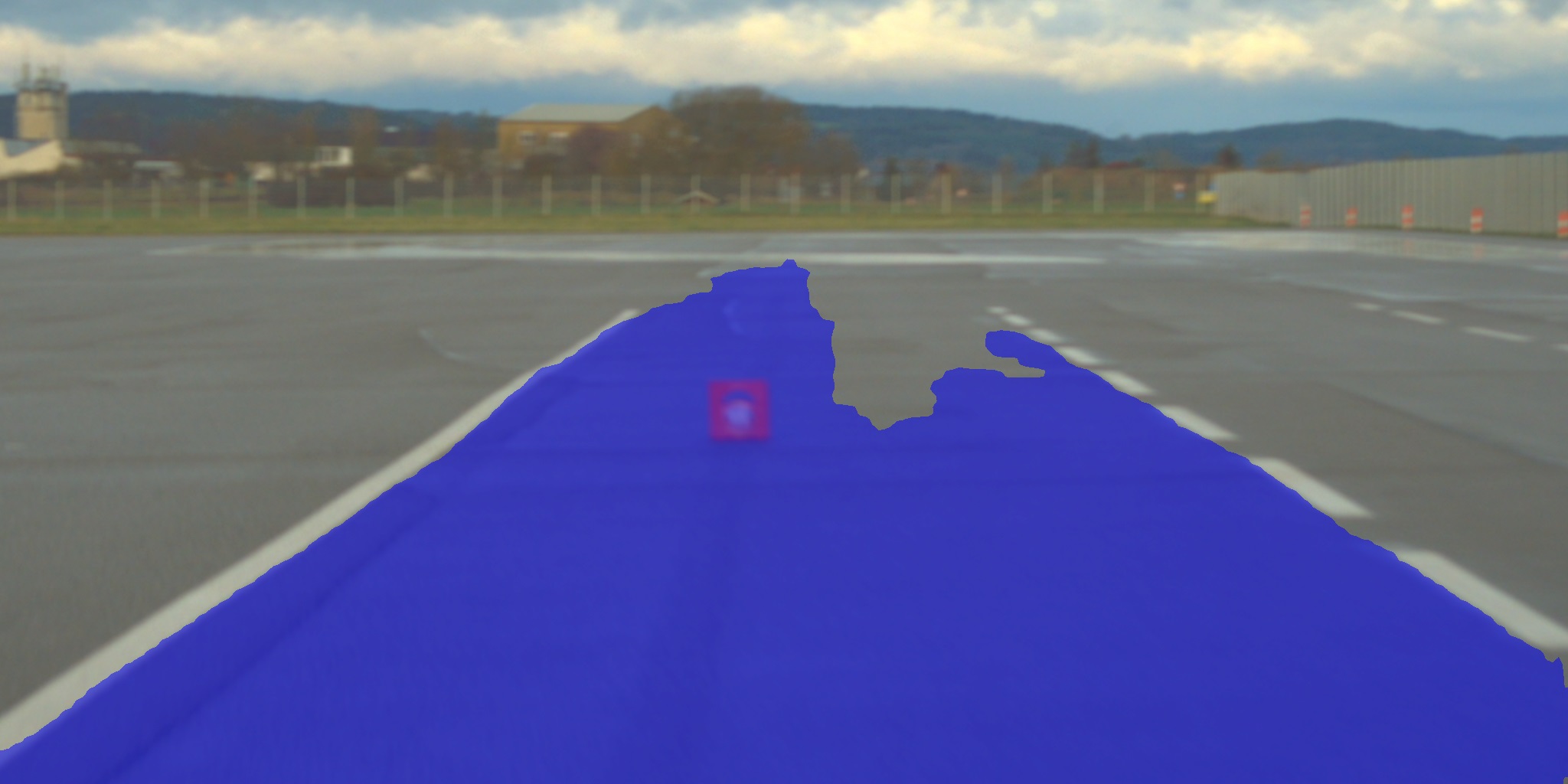}}
		\label{fig:pic1}
	\end{subfigure}
	\begin{subfigure}{.2\textwidth}
		\centering
		\small\textbf{Energy Map}\par\medskip
		\setlength{\fboxsep}{0pt}\fbox{\includegraphics[width=.99\linewidth]{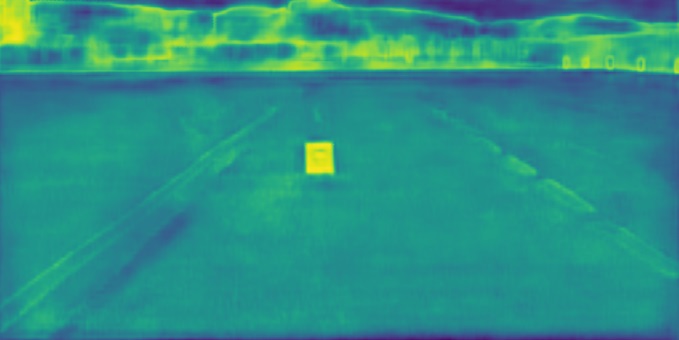}}
		\label{fig:pic2}
	\end{subfigure}
	\begin{subfigure}{.2\textwidth}
		\centering
		\small\textbf{Outlier Pixels}\par\medskip
		\setlength{\fboxsep}{0pt}\fbox{\includegraphics[width=.99\linewidth]{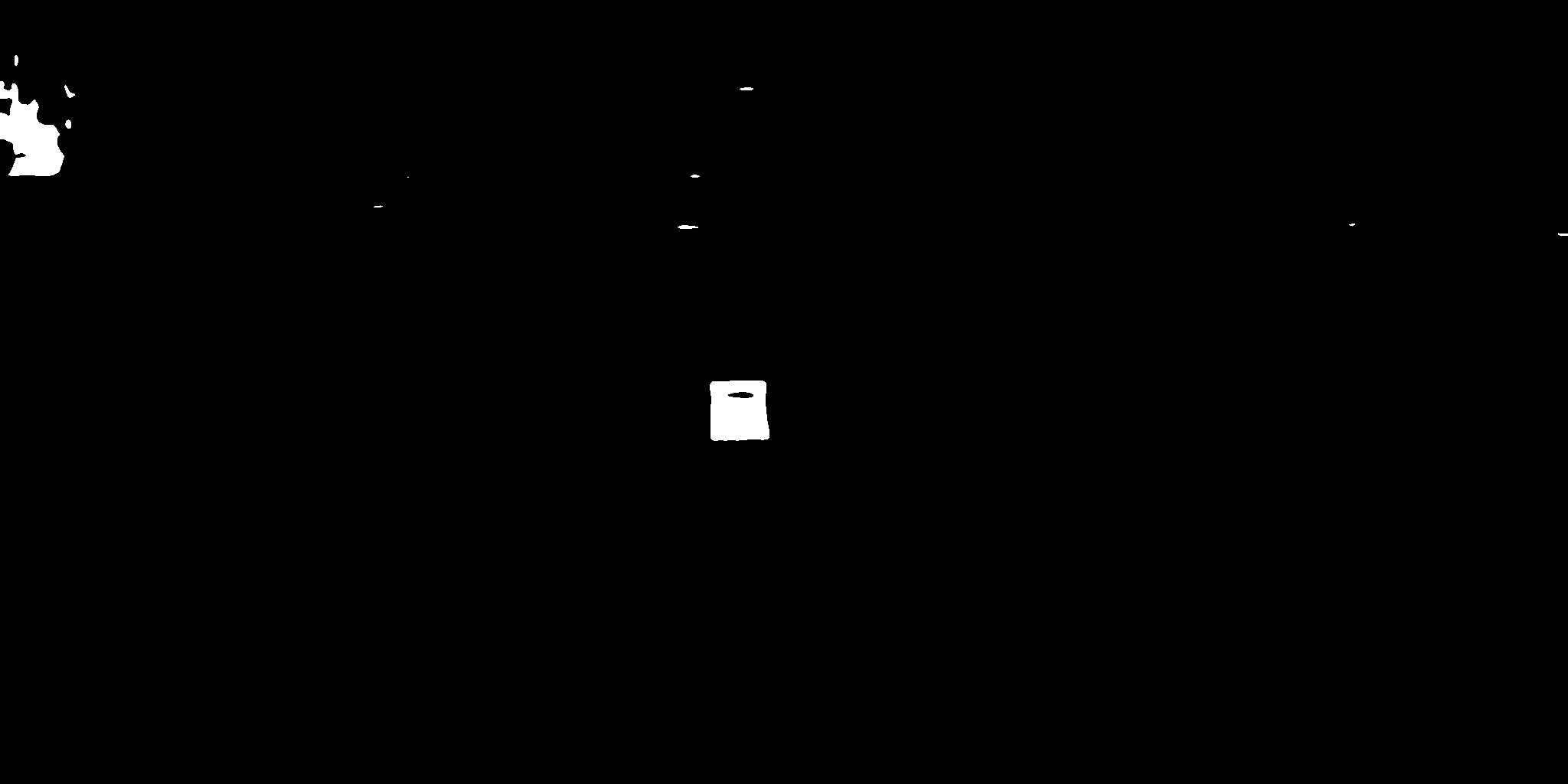}}
		\label{fig:pic3}
	\end{subfigure}
	\begin{subfigure}{.2\textwidth}
		\centering
		\small\textbf{Final Ego-Corridor}\par\medskip
		\setlength{\fboxsep}{0pt}\fbox{\includegraphics[width=.99\linewidth]{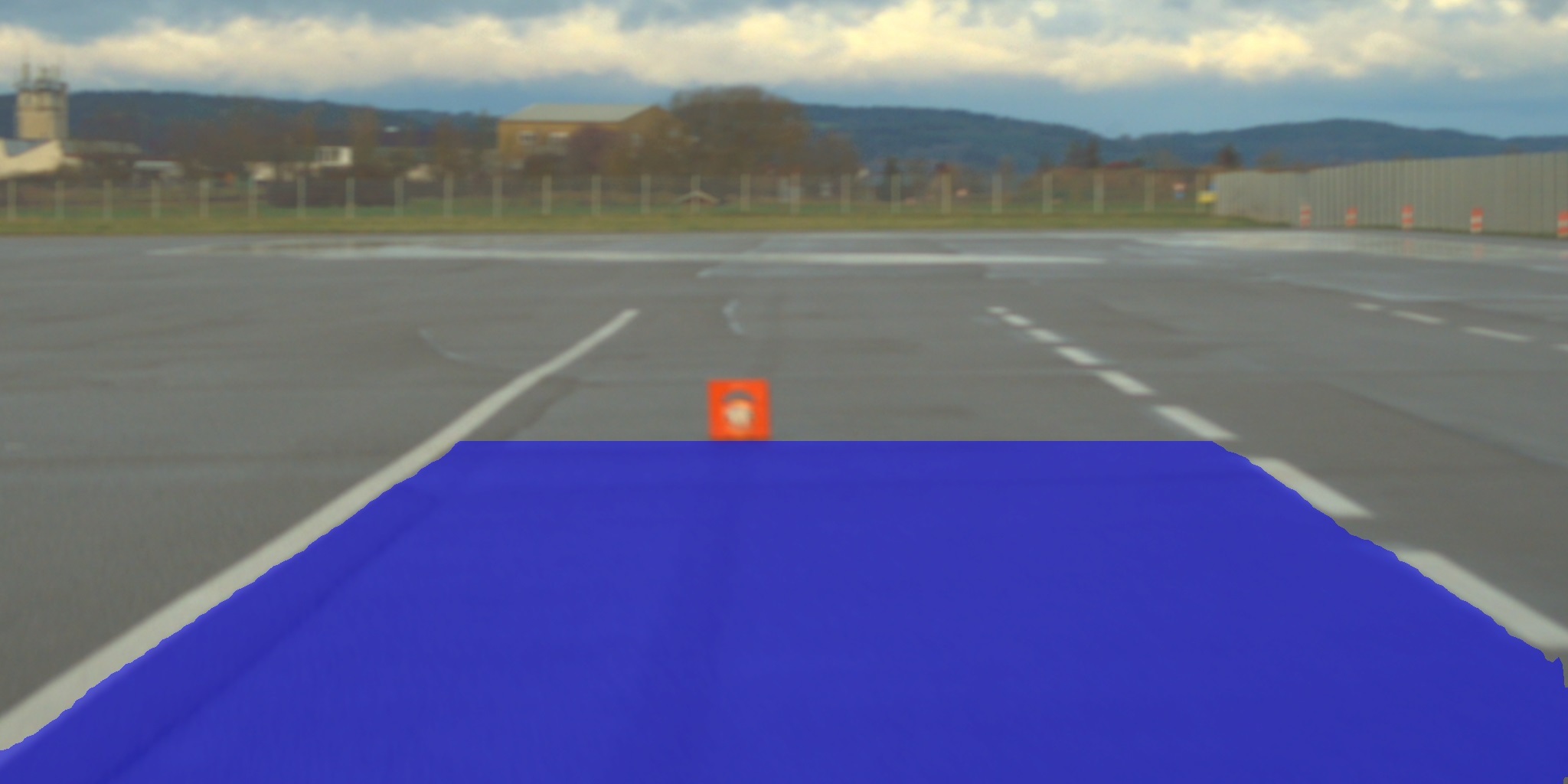}}
		\label{fig:pic4}
	\end{subfigure}
	\vspace{-2ex}
	\begin{subfigure}{.2\textwidth}
		\centering
		\setlength{\fboxsep}{0pt}\fbox{\includegraphics[width=.99\linewidth]{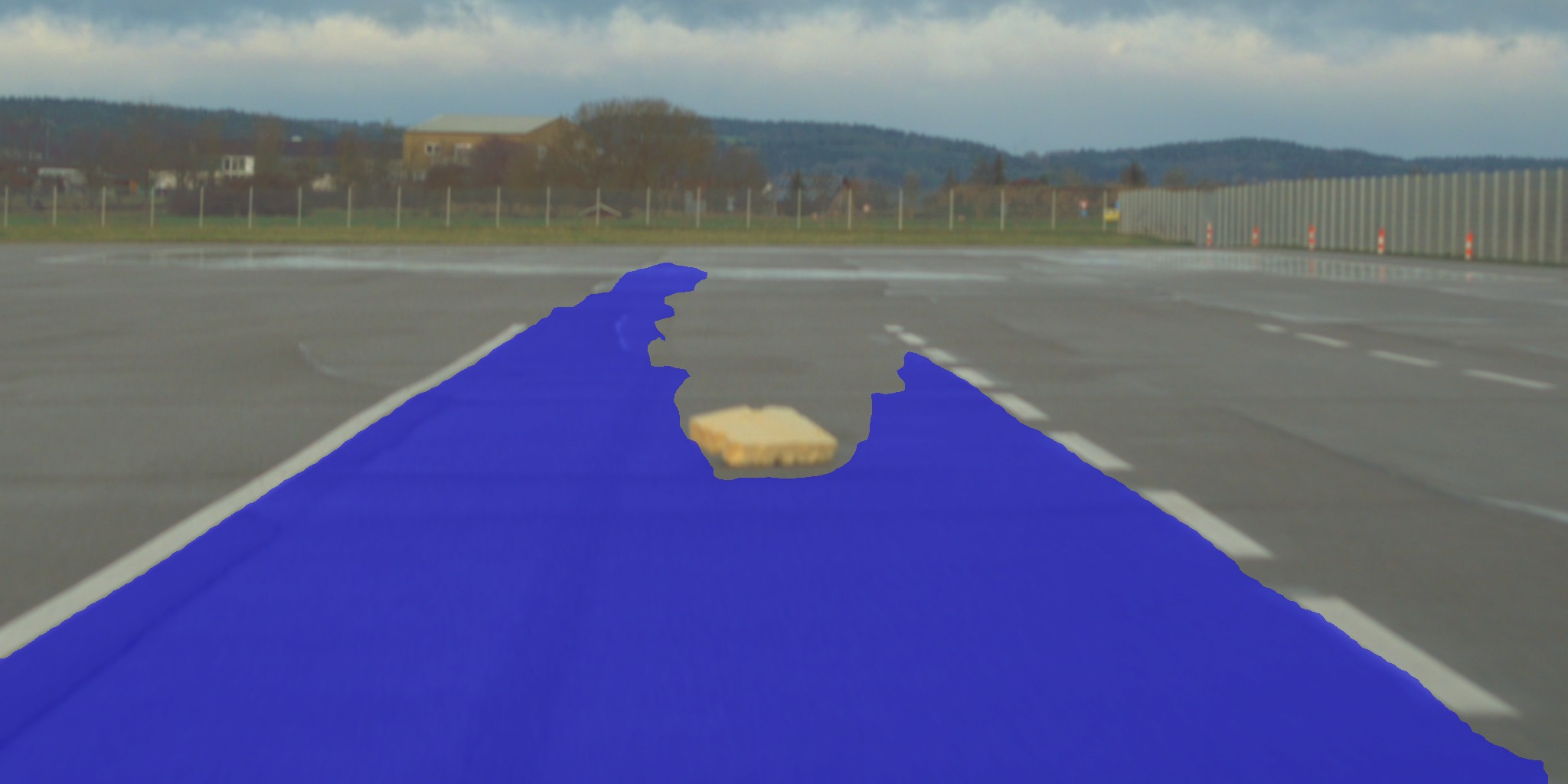}}
		\label{fig:pic1}
	\end{subfigure}
	\begin{subfigure}{.2\textwidth}
		\centering
		\setlength{\fboxsep}{0pt}\fbox{\includegraphics[width=.99\linewidth]{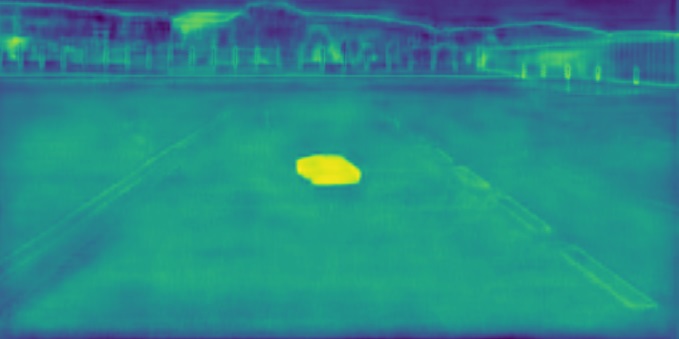}}
		\label{fig:pic2}
	\end{subfigure}
	\begin{subfigure}{.2\textwidth}
		\centering
		\setlength{\fboxsep}{0pt}\fbox{\includegraphics[width=.99\linewidth]{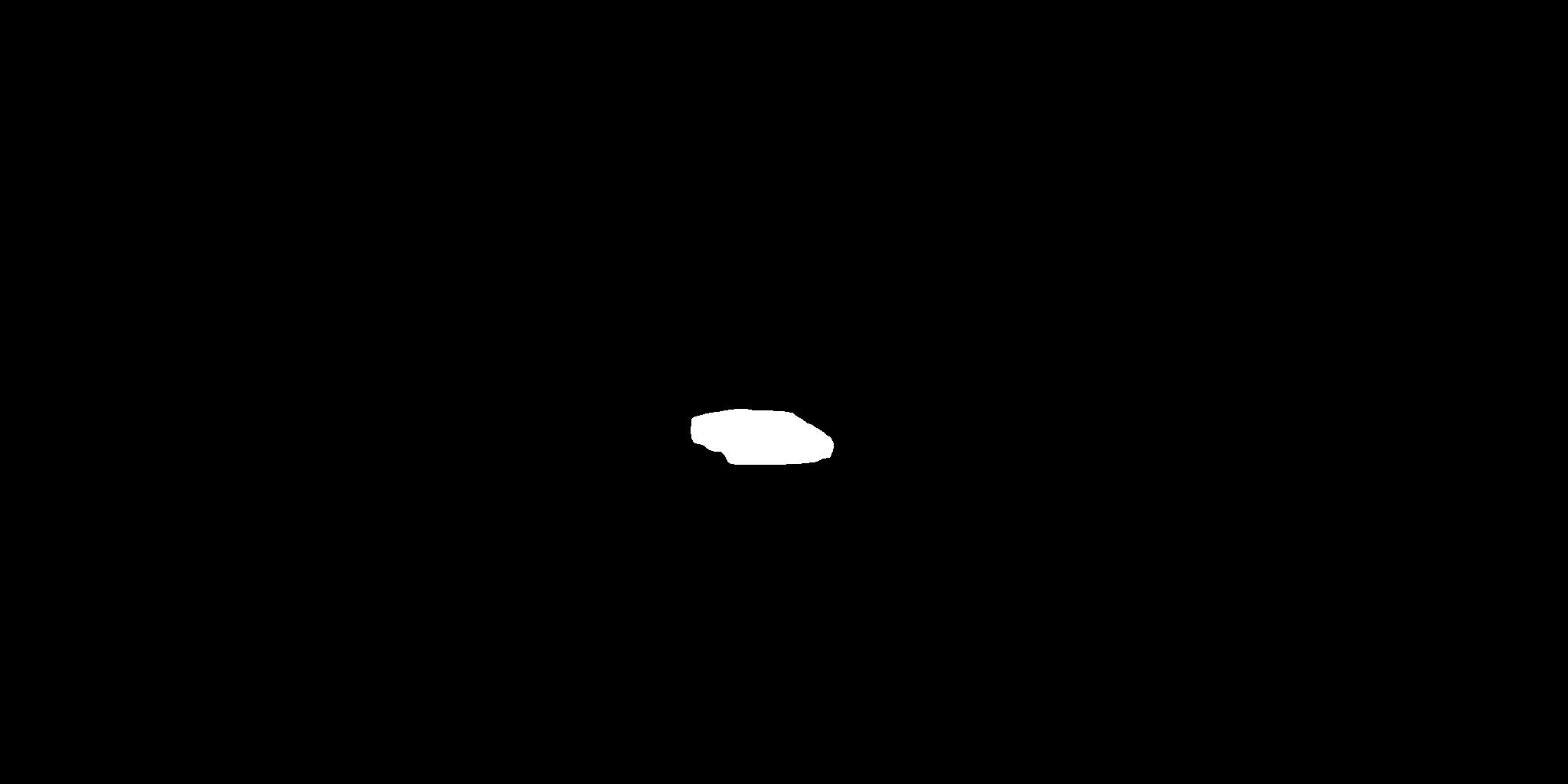}}
		\label{fig:pic3}
	\end{subfigure}
	\begin{subfigure}{.2\textwidth}
		\centering
		\setlength{\fboxsep}{0pt}\fbox{\includegraphics[width=.99\linewidth]{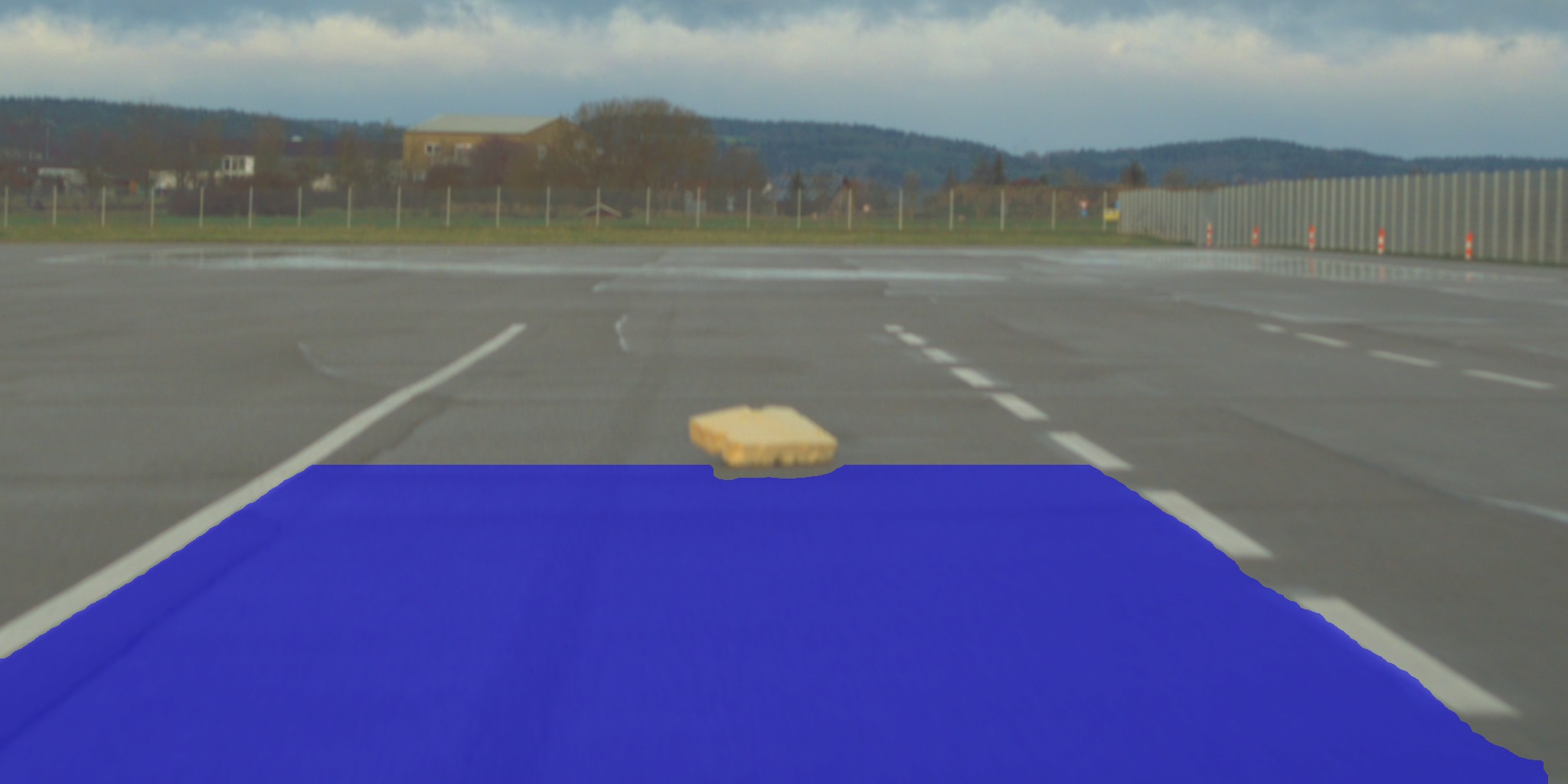}}
		\label{fig:pic4}
	\end{subfigure}
        \caption{Qualitative results for handling occasionally missing sharp cut-offs at closer distances: (a) Ego-corridor path, (b) Energy map of PEBAL path, (c) Detected outlier pixels, (d) Final ego-corridor as a result from combing (a) and (c).}
        \vspace{-0.4cm}
	\label{fig:mtl_results}
\end{figure*}
\subsection{Quantitative Results}
Table \ref{table:lostcargo_results} shows the quantitative results for our ego-corridor-only architecture trained with the three different approaches, as well as for our multitask approach.
Looking at the results, several interesting findings can be noticed:
Firstly, all ego-corridor only models perform significantly better for objects at large distances compared to near objects. At first glance, this seems very counter-intuitive. Objects at closer distance appear significantly larger and more refined than at far distances. As such, it should theoretically be easier to detect them and create the correct ego-lane accordingly.

And, indeed, oftentimes, the models understand at close distance, that an object is present. In these cases, the issue is not the detection of the object itself, but the correct longitudinal limitation of the ego-lane. Figure \ref{fig:mtl_results} exemplarily shows the results of the multitask network and the output of the two combined paths (ego-corridor path and PEBAL path) for objects at a distance of 25\,m, where the ego-corridor clearly detects the objects, but is unable to use that knowledge to further create the correct ego-corridor segmentation (sharp cut-off). Instead, the objects are merely circumvented and the ego-corridor continues well past the objects. We use the simple post-processing described in Chapter \ref{chap:methods}, but a more elaborate approach could easily increase performance in these situations.

Secondly, the model trained using \textit{Naive} training scheme that has no direct notion of non-traffic related obstacles, performs very well for larger distances detecting more than $80\,\%$ of the objects at distances larger than 200\,m. On the contrary, the \textit{synthetic object training} procedure achieves the worst performance by far for objects at a large distance and in general does not perform better than the \textit{Obstacle} training procedure. This suggests that the model cannot translate the detection of synthetic objects to the detection of real objects very well. So, the domain shift unfortunately over-compensates the gain from more and diverse data.

Thirdly, the multitask approach sees a performance increase in the near range compared to the ego-corridor-only approach but also significantly reduces performance in the far range. This issue is caused by noise in the far range that is falsely detected as outliers and causing the ego-corridor to thus be wrongly limited. For the near range, however, the method performs very well, achieving the best result of all of our approaches. The unexpected obstacle path is able to detect many of the objects at this distance and using this information the model is able to create the ego-corridor mask correctly, where it would have over-segmented the object without the explicit information.
\subsection{Qualitative Results}
Figure \ref{fig:obst_qualitative_res} shows exemplary ego-corridor predictions for an object at a distance of 300\,m as well as a comparison between predictions with and without post-processing. Fig. \ref{fig:mtl_results} shows the intermediate outputs as well as the final output of the multitask network.

As the figures show, even at a large distance of 300\,m and an object that consists of very few pixels in the image, the ego-lane is correctly segmented. We can also see that the simplistic post-processing procedure is very successful in producing a sharp edge in front of an object, which was correctly detected but not correctly limiting the ego-lane without the post-processing.
In order to further ease understanding of the benefits of this approach, we provided a video \cite{Video_LostCargo} showing one representative sequence (test data) with all intermediate and final outputs (ego vehicle driving 120 kph). The approach was successfully tested on a prototype vehicle and was able to reach real-time on 10Hz input images.

The multitask outputs show that the method has potential of achieving good results in explicitly detecting the objects in one path, creating a good approximation of the ego-corridor in the other path and combining this information to create an accurate ego-corridor even when the ego-corridor path misses the object.
%tly at 25\,m, it also contains a lot of noise. As the TSC images differ significantly in appearance, the model sometimes does not at all recognize the scene, classifying large parts of the image as outliers.

	%
	\section{Conclusion and Outlook}
	Unexpected obstacle detection is a challenging task. We have shown in this
	work, that an implicit approach with a data-based ego-corridor rather than an explicit detection of these obstacles can achieve very good results.
	Paired with a tele mono video-camera, the approach achieves a detection rate of $95\,\%$ for out-of-distribution objects at 300\,m. This far surpasses the range and quality of the original ego-corridor as well as the known
	explicit obstacle detection methods with similar run time complexity.
	We have also shown that an additional cheap explicit detection method can support the ego-corridor in those cases that it misses an obstacle, especially so in the near range, 	where the ego-corridor is biased towards expecting an intact drivable space.

	In the future, we want to train the PEBAL path of the multitask approach on the same dataset as the ego-corridor path so that the entire network can be trained end-to-end at the same time and no transfer learning from Cityscapes is necessary. We further want to look into more sophisticated post-processing approaches that could significantly improve the ego-corridor performance in cases where the objects are clearly detected but the ego-corridor was not shaped accordingly. More online testing including realistic scenarios with lost objects in inner city will be done to ensure operability in open-world context. Also more conceptual work might be required to be able to differentiate uncritical cases (e.g. road ends at a T-junction) from road blockages. Our still to be proofed assumtion would be that other system layers (e.g., the behavior planner) will handle such cases taking additional information as map data into account.

	\addtolength{\textheight}{-12cm}   % This command serves to balance the column lengths

	%%%%%%%%%%%%%%%%%%%%%%%%%%%%%%%%%%%%%%%%%%%%%%%%%%%%%%%%%%%%%%%%%%%%%%%%%%%%%%%%
	%References are important to the reader; therefore, each citation must be complete and correct. If at all possible, references should be commonly available publications.
	%working for git version
	\bibliographystyle{./bibtex/IEEEtran} % use IEEEtran.bst style
	\bibliography{./bibtex/IEEEabrv,./bibtex/paper}
\end{document}